\documentclass{article}

\usepackage[preprint, nonatbib]{neurips_2024}

\usepackage[utf8]{inputenc} % allow utf-8 input
\usepackage[T1]{fontenc}    % use 8-bit T1 fonts
\usepackage{hyperref}       % hyperlinks
\usepackage{url}            % simple URL typesetting
\usepackage{amsfonts}       % blackboard math symbols
\usepackage{nicefrac}       % compact symbols for 1/2, etc.
\usepackage{microtype}      % microtypography
\usepackage[svgnames]{xcolor}         % colors
\usepackage{amsmath}
\usepackage{enumitem}
\usepackage{amsthm}
\usepackage{subfigure}
\usepackage{float}
\usepackage{makecell}
\usepackage{natbib}
\usepackage{soul}
\usepackage{graphicx}
\usepackage{caption}    % 主 caption 支持
\usepackage{subcaption}
\usepackage{booktabs}       % professional-quality tables
\usepackage{multirow}
\usepackage[most]{tcolorbox}
\usepackage{textcomp}
\usepackage{xurl}

\newtheorem{proposition}{Proposition}  % 定义“命题”环境

\newcommand{\mrc}[1]{\textcolor{red}{#1}}

\newcommand{\rit}[1]{\textcolor{blue}{#1}}

\title{Good Learners Think Their Thinking: Generative PRM Makes Large Reasoning Model More Efficient Math Learner}

\author{
    Tao He\textsuperscript{1}\thanks{\quad Equal contribution.} \quad 
    Rongchuan Mu\textsuperscript{1}\footnotemark[1]\quad
    Lizi Liao\textsuperscript{2}\quad
    Yixin Cao\textsuperscript{3}\quad
    \textbf{Ming Liu}\textsuperscript{\textbf{1}} \hspace{-1mm}\footnotemark[2]\thanks{\quad Corresponding author.} \quad \textbf{Bing Qin}\textsuperscript{\textbf{1}}\\
    \textsuperscript{1} Harbin Institute of Technology \\
    \textsuperscript{2} Singapore Management University \\
    \textsuperscript{3} Fudan University \\
   Corresponding to: \texttt{the@ir.hit.edu.cn}
}

\begin{document}

\maketitle

\begin{abstract}
%This study investigates the application of process rewards to accelerate the training efficiency of large reasoning models (LRMs) in reinforcement learning. Traditional process reward models (PRMs) often face challenges such as high reasoning capacity requirements, unclear step granularity, and susceptibility to reward hacking. To overcome these issues, we propose a novel comprehension-driven Generative PRM that operates at the thought level. This design enables more reliable and structured credit assignment, alleviates ambiguity in step segmentation, and reduces reliance on the reward model’s own reasoning capabilities by only primarily leveraging its comprehension strength.
%In addition, we introduce an adaptive reward mechanism that dynamically aligns with the LRM’s current reasoning capacity, effectively balancing exploration and exploitation. Building on the evaluated process rewards, we develop a new reinforcement learning algorithm, TP-GRRO, implemented by an efficient off-policy training pipeline to reduce GPU idling.
%Experiments on the DeepSeek-R1-Distill-Qwen model family demonstrate that our approach consistently outperforms conventional outcome-reward-based training paradigms on mathematical reasoning tasks, validating the core hypothesis that well-structured process rewards can substantially accelerate LLM learning.

Large reasoning models (LRMs) have recently shown promise in solving complex math problems when optimized with Reinforcement Learning (RL). But conventional approaches rely on outcome-only rewards that provide sparse feedback, resulting in inefficient optimization process. 
% We propose a novel comprehension-driven generative process reward model (\textbf{GenPRM}) operating at the thought level to address major bottlenecks in reasoning reinforcement learning. 
In this work, we investigate the function of process reward models (PRMs) to accelerate the RL training for LRMs.
We propose a novel intrinsic signal-driven generative process evaluation mechanism operating at the thought level to address major bottlenecks in RL-based training. 
Specifically, instead of requiring PRMs to know how to solve problems, our method uses intrinsic signals in solutions to judge stepwise correctness and aggregate contiguous correct/incorrect steps into coherent `thought' units. 
% We propose a novel intrinsic signal-driven generative process evaluation mechanism operating at the thought level to address major bottlenecks in RL-based training. 
% % Instead of requiring the reward model to fully solve problems, our GenPRM uses basic comprehension to judge stepwise correctness, aggregating contiguous correct or incorrect steps into coherent `thought' units. 
% Instead of requiring the PRMs to know how to solve problems, our method uses basic comprehension to judge stepwise correctness, aggregating contiguous correct/incorrect steps into coherent `thought' units. 
This structured, thought-level rewards enable more reliable credit assignment by reducing ambiguity in step segmentation and alleviating reward hacking. 
We further introduce a capability-adaptive reward mechanism that dynamically balances exploration and exploitation based on the LRM’s current proficiency, guiding learning without stifling creative trial-and-error. 
% These innovations are integrated into a new off-policy RL algorithm, \textbf{TP-GRPO}, which extends Grouped Proximal Optimization with process-based rewards while improving training efficiency by reusing trajectories. 
These innovations are integrated into a new off-policy RL algorithm, \textbf{TP-GRPO}, which extends grouped proximal optimization with process-based rewards and improves training efficiency. 
% Experiments on 1.5B and 7B parameter LRMs math solvers demonstrate that our method achieves higher problem-solving accuracy with significantly fewer training samples than outcome-only reward baselines. 
Experiments on 1.5B and 7B parameter LRMs demonstrate that our method achieves higher problem-solving accuracy with significantly fewer training samples than outcome-only reward baselines. 
% The results validate that well-structured process rewards can substantially accelerate LLM learning on challenging reasoning tasks.
The results validate that well-structured process rewards can substantially accelerate LRM optimization in math reasoning tasks. Code is available at  ~\url{https://github.com/cs-holder/tp_grpo}.

\end{abstract}

\section{Introduction}

% The remarkable success of outcome reward approaches, exemplified by DeepSeek R1 and Kimi-k1.5, has demonstrated the effectiveness of Reinforcement Learning with Verified Rewards (RLVR). The core of RLVR lies in the application of efficient rule-based outcome reward functions, which enable Reinforcement Learning (RL) scaling.
The remarkable success of Reinforcement Learning with Verifiable Rewards (RLVR) methods~\citep{guo2025deepseek, liu2025understanding, hu2025open}, exemplified by DeepSeek R1~\citep{guo2025deepseek} and Kimi k1.5~\citep{team2025kimi}, demonstrates the effectiveness of Reinforcement Learning (RL) in enhancing the reasoning capabilities of Large Language Models (LLMs). At their core, these methods scale RL by employing efficient and model-free rule-based outcome reward functions~\citep{shao2024deepseekmath, yue2025does, xu2025direct}.
% and have shifted the research focus toward data construction, reward function design, and adaptive reasoning. 
% It is well-known that the balance between exploration and exploitation poses the key topic of RL. RLVR encourages broad exploration of the state space via efficient reward functions—akin to students improving through repetitive practice, yet neglecting reflection on their solution strategies. In contrast, efficient learners reflect on their reasoning paths after each practice session, thereby accelerating knowledge acquisition. 
% In contrast, the modeling of process rewards has been largely marginalized. 
% This analogy motivates a fundamental question: can process-level reward mechanisms guide Large Reasoning Models (LRMs) to engage in selective learning along reasoning trajectories, ultimately improving training efficiency and sample utilization?
It is well established that RL revolves around two fundamental challenges: exploration and exploitation~\citep{berger2014exploration, wang2018exploration, fruit2019exploration, hao2025llm}. RLVR-style approaches exploration by leveraging efficient reward functions to enable large-scale exploration, akin to how learners improve problem-solving skills through extensive practice without retrospect their thinking processes.

% \dif{It is well established that RL fundamentally centers on the trade-off between exploration and exploitation~\citep{berger2014exploration, fruit2019exploration, hao2025llm}. 
% RLVR-style approaches exploration by leveraging efficient outcome reward functions, akin to how students improve problem-solving skills through extensive practice without reflection.}

However, such methods focus solely on the correctness of final answers and neglect to exploit the informative value of intermediate reasoning trajectories~\citep{chen2025reasongrm, zou2025reasonflux, chen2024step, ye2025uncertainty}. 
% \mrc{In contrast, skilled learners reflect on their reasoning process after each attempt, engaging in selective learning to maximize the learning value of each exercise.
% Analogously, relying only on outcome rewards in math reasoning for Large Reasoning Models (LRMs) may similarly limit their learning efficiency. Motivated by this insight, we explore a fundamental question: \textbf{whether introducing process-level rewards can guide LRMs to better exploitation and selectively learn from their reasoning trajectories, thereby improving the overall training efficiency.} }
In contrast, skilled learners reflect on their redundant thinking after exercise, engaging in selective learning to maximize the learning efficiency. 
This strategy inspires the use of process rewards to guide the learning of Language Reasoning Models (LRMs). 
Motivated by this analogy, we want to explore a central question: \textbf{can well-designed process rewards enhance the learning efficiency of LRMs?} 
% Building upon this investigation, we further revisit the study value of process rewards in this era of RLVR.
We hope to motivate a broader discussion on the significance of process rewards for LRM training.

Prior to DeepSeek R1~\citep{guo2025deepseek}, learning discriminative process reward models (PRMs) had already received considerable attention. 
OpenAI released a public dataset for process evaluation~\citep{lightman2023let}; \cite{wang2023math, wang2024multi} proposed a Monte Carlo Tree Search-based method to automatically annotate process rewards; and Qwen’s technical report explored the development of Process Reward Models~\citep{zhang2025lessons} and offered a benchmark for measuring the ability to assess the quality of the reasoning trace~\citep{zheng2024processbench}.
% \dif{
% OpenAI introduced a public dataset for process evaluation~\citep{lightman2023let}; 
% % \citet{wang2023math} proposed a Monte Carlo Tree Search-based approach for automatic process reward annotation; 
% % \citet{wang2023math} proposed an automatic approach for process reward annotation; 
% and Qwen's team also explored the bitter lesson of Process Reward Models (PRMs)~\citep{zhang2025lessons}. 
% Nonetheless, achieving reliable process evaluation remains a significant challenge.
% }
Nevertheless, DeepSeek R1 revealed the limitations of discriminative PRMs, including difficulty in step segmentation, limited generalization, vulnerability to reward hacking, and training inefficiencies~\citep{guo2025deepseek}. 
% As a result, recent efforts have shifted toward Generative Process Reward Models (GenPRMs), which leverage powerful reasoning models to generate process evaluations with detailed reasoning processes.
Therefore, some recent efforts have shifted toward Generative PRMs (GenPRMs)~\citep{khalifa2025process, feng2025prm, she2025r}, 
% \mrc{which leverage powerful reasoning models to evaluate reasoning trajectories with detailed reasoning processes, in contrast to traditional discriminative reward modeling~\citep{liu2025inference, chen2025rm}.}
which leverage powerful LRMs to evaluate reasoning processes with detailed thinking~\citep{liu2025inference, chen2025rm}.

In light of existing critiques of PRMs~\citep{guo2025deepseek, song2025prmbench, processbench}, we argue that a robust GenPRM-based process evaluation mechanism should satisfy three essential criteria:
\textbf{1) Appropriate unit segmentation.} Excessively fine-grained evaluation increases the burden of process evaluation, thus amplifying the risk of flawed evaluation. 
% And excessive coarseness in segmentation can obscure the boundary between correct and incorrect reasoning steps.
And excessive coarseness in segmentation can also result in correct and incorrect steps being grouped together.
% , thereby impairing evaluation accuracy.
\textbf{2) Evaluation capability decoupled from reasoning proficiency.} Current GenPRM approaches often assess step correctness via thinking mode~\citep{feng2025prm}, which implicitly assumes the PRM can solve the original problem. This strong prerequisite elevates the reasoning ability demands on the GenPRM, limiting its reliability when facing difficult questions.
\textbf{3) Balanced exploration and exploitation.} 
% The incorporation of process rewards should strike an effective balance between exploration and exploitation: enhancing trajectory exploitation while preserving the exploration ability of LRMs.
Overemphasizing process rewards may impair the LRM’s exploratory capabilities, like trial-and-error reasoning, reflection, and self-correction. 
Therefore, it is crucial to mitigate the negative impact of process rewards on exploration while leveraging their benefits for guided learning. 

Grounded in these criteria, we design a new GenPRM-based process evaluation mechanism, incorporating three principal innovations compared to existing methods:
\textbf{1)} We propose a thought-level evaluation strategy that aggregates contiguous correct or incorrect reasoning steps into coherent evaluation units, termed thoughts.
% and assigns process rewards at this higher level of abstraction. 
This design anchors reward assignment in the correctness rather than the semantic granularity of step segmentation, thereby reducing sensitivity to inconsistent step definitions.
\textbf{2)} We introduce an intrinsic signal-driven evaluation paradigm that significantly lowers the reliance on the reasoning capabilities of the GenPRM. 
% Unlike prior GenPRMs that require strong multi-step reasoning abilities, our approach assumes only basic instruction-following and process comprehension, thus relaxing the capacity demands on the reward model.
Unlike prior GenPRMs that require strong multi-step reasoning abilities, our method leverages intrinsic signals within reasoning trajectories for process evaluation, relying primarily on comprehension of the reasoning process, thus relaxing the capability requirements on the reward model.
\textbf{3)} We develop a capability-adaptive reward computation mechanism that dynamically calibrates reward signals based on the current reasoning proficiency of the LRM. This mechanism facilitates a better balance between exploration and exploitation during RL training, while maintaining focus on critical segments relevant to the final answer. Crucially, it enables guided learning with greater robustness against reward hacking.
Built upon this enhanced process evaluation framework, we modify the GRPO~\citep{shao2024deepseekmath} and propose our TP-GRPO (\textbf{T}hought-level \textbf{P}rocess-reward-based \textbf{GRPO}) algorithm. 
Considering the inefficiencies of GenPRM, we further design an efficient off-policy training pipeline that mitigates GPU idling caused by the generative process evaluation.

% We validate our approach on the DeepSeek-R1-Distill series of models. Our 1.5B model, trained with only 700 problems, achieves a +4.7 Pass@1 improvement on AIME 2024; with 1,800 training samples, the improvement reaches +5.98, significantly outperforming outcome-reward-only baselines. These results strongly support our central hypothesis: judicious integration of process rewards can accelerate complex reasoning learning.
We validate our approach on the DeepSeek-R1-Distill-Qwen family of models~\citep{guo2025deepseek}. Our 1.5B model, trained on only 700 problems, achieves +4.32\% improvements in Pass@1 on the AIME 2024 benchmark~\citep{AIME24}; with 1800 training problems, the improvements increase to +5.98\%. Remarkably, the 7B model, trained on just 1070 problems, also achieves +6.67\% gain in Pass@1, substantially outperforming baselines trained with outcome-only rewards. These results strongly support our central hypothesis: that the judicious introduction of process-level rewards can significantly accelerate learning in math reasoning tasks.

Our main contributions are summarized as follows:
% \begin{itemize}
%     \item We propose a new Generative PRM approach that drastically reduces reliance on reasoning capabilities. We also introduce a thought-level process reward modeling paradigm that addresses challenges such as segmentation granularity and reward hacking.
%     \item We design a difficulty-adaptive process reward mechanism, and integrate it with GRPO to construct an improved RL paradigm that effectively balances exploration and exploitation. We further introduce an efficient off-policy training pipeline that substantially enhances training throughput.
%     \item We conduct comprehensive experiments on both 1.5B and 7B LRM scales, with performance gains validating the effectiveness and adaptability of our proposed method. Besides, study results provide preliminary evidence supporting our hypothesis: incorporating process rewards significantly improves the learning efficiency, particularly for Qwen 1.5 and 7B models, compared to using only outcome rewards.
% \end{itemize}
\begin{itemize}
    \item We propose a new GenPRM paradigm that advances existing approaches along three dimensions: evaluation granularity, dependency on reasoning capabilities, and reward formulation. Our design addresses key limitations such as ambiguous process segmentation, over-reliance on reasoning strength, and reward hacking.

    % \dif{
    % We propose a new GenPRM paradigm to address key limitations such as ambiguous process segmentation, over-reliance on reasoning strength, and reward hacking.
    % }

    \item We integrate the proposed process reward mechanism with GRPO and introduce a novel RL algorithm, TP-GRPO. To address the efficiency bottlenecks introduced by GenPRM, we further develop a scalable off-policy training pipeline.

    % \dif{
    % , as well as an efficient off-policy training pipeline.
    % }

    % \item \mrc{Experiments on DeepSeek-R1-Distill 1.5B and 7B models demonstrate the effectiveness and adaptability of our method. The observed performance gains provide empirical support for our central hypothesis: incorporating reasonable process-level rewards significantly improves learning efficiency compared to using outcome-only rewards, especially on strong LRMs such as DeepSeek-R1-Distill series models.}

    \item Experiments on DeepSeek-R1-Distill 1.5B and 7B models demonstrate the effectiveness and efficiency of our approach. The performance gains empirically support our core hypothesis: integrating well-designed process rewards significantly enhances learning efficiency over outcome-only supervision GRPO.
\end{itemize}

\section{Preliminary}\label{sec:preliminary}
\subsection{RL Foundation: GRPO}\label{sec:grpo_foundation}
Our method builds on GRPO, which estimates advantage values via Monte Carlo (MC) sampling, thereby avoiding eliminating the need to learn a value function. Specifically, given a query $q$, GRPO samples $G$ solutions ${o_i}_{i=1}^G$ from the old policy $\pi_{\theta_{old}}$ and optimizes the current policy $\pi_\theta$ using:
% \begin{equation}
%     \begin{aligned}
%         \mathcal{J}_{GRPO}(\theta) &= \mathbb{E}[{q\sim P(Q), \{o_i\}\sim\pi_{\theta_{old}}(Q|q)}]\\
%         &\frac{1}{G}\sum_{i=1}^G\frac{1}{|o_i|}\sum_{t=1}^{|o_i|} \left\{ \min\left[\rho_{i,t} \hat{A}_{i,t}, \text{clip}(\rho_{i,t}, 1-\epsilon, 1+\epsilon)\hat{A}_{i,t}\right]  - \beta D_{\mathrm{KL}}[\pi_\theta \| \pi_{\mathrm{ref}}] \right\},
%     \end{aligned}
% \end{equation}
\begin{equation}
    \begin{aligned}
        \mathcal{J}_{\text{GRPO}}(\theta) &= \mathbb{E}[{q\sim P(Q), \{o_i\}_{i=1}^G\sim\pi_{\theta_{old}}(Q|q)}]\\
        &\frac{1}{\sum_{i=1}^G|o_i|}\sum_{i=1}^G\sum_{t=1}^{|o_i|} \left\{ \min\left[\rho_{i,t} \hat{A}_{i,t}, \text{clip}(\rho_{i,t}, 1-\epsilon_{\text{low}}, 1+\epsilon_{\text{high}})\hat{A}_{i,t}\right]  - \beta D_{\mathrm{KL}}[\pi_\theta \| \pi_{\mathrm{ref}}] \right\},
    \end{aligned}\label{eq:grpo_loss}
\end{equation}
where $\rho_{i,t} = \frac{\pi_{\theta}(o_{i,t} \mid q, o_{i,<t})}{\pi_{\theta_{old}}(o_{i,t} \mid q, o_{i,<t})}$ refers to the policy ratio, and $\hat{A}_{i,t}$ denotes the estimated advantage for the $t$-th token of solution $o_i$. Here, we adopt two effective improvements from DAPO over GRPO, including the use of token-level loss instead of solution-level loss, as well as the application of distinct clipping ranges, $ \epsilon_{\text{low}} $ and $ \epsilon_{\text{high}} $. Depending on the availability of process-level rewards, GRPO supports two variants of advantage estimation:
\begin{itemize}[leftmargin=1cm]
    \item \textbf{With Outcome-only Rewards.} Motivated by the dominant performance of DeepSeek R1, most RLVR research works rely solely on rule-based outcome rewards. For each sampled solution $o_i$, a scalar outcome reward $r_i^o$ is assigned. Rewards in a group are normalized as:
    \begin{equation}
        \hat{r}_i^o = \frac{r_i^o - \text{mean}(\{r_j^o\}_{j=1}^G)}{\text{std}(\{r_j^o\}_{j=1}^G)}
        \label{eq:reward_normalize}
    \end{equation}
    In this case, the advantage of each token is set uniformly as $A_{i,t} = \hat{r}_i^o$.

    \item \textbf{With Process Rewards.} When intermediate step-level rewards are accessible, the computation of the advantage differs accordingly. 
    Let $R = \{{r_1^{\text{index}(1)}, \dots, r_1^{\text{index}(K_1)}}, \dots, $ ${r_G^{\text{index}(1)}, \dots, r_G^{\text{index}(K_G)}}\}$ denote the full set of rewards across all $G$ solutions, where $\text{index}(k)$ denotes the position of the last token in the $k$-th step.
    GRPO first normalizes all step rewards:
    \begin{equation}
        \hat{r}_i^{\text{index}(j)} = \frac{r_i^{\text{index}(j)} - \text{mean}(R)}{\text{std}(R)}
    \end{equation}
    and then computes the advantage of token $t$ in $o_i$ as the cumulative sum of normalized rewards from all steps ending at or after the $t$-th token:
    \begin{equation}
        A_{i,t} = \sum_{\text{index}(j) \ge t} \hat{r}_i^{\text{index}(j)}
        \label{eq:process_reward_advantage}
    \end{equation}
\end{itemize}

Previous attempts to utilize process rewards often relied on learning a specialized process reward model, which suffers from challenges like reduced accuracy and vulnerability to reward hacking. As a result, interest in process rewards declined after the emergence of strong outcome-based baselines.
In contrast, our work avoids explicitly training the process reward model. Instead, we leverage the outstanding reasoning capability of LRM to evaluate intermediate steps.
% , aiming to improve evaluation quality while mitigating reward hacking risks.

\subsection{Problem Formulation}
Given a math problem $x$, the LRM $\pi_\theta$ generates a Long CoT $o$, which typically consists of two parts: \textit{\textbf{think}} and \textit{\textbf{answer}}~\cite{guo2025deepseek}. 
The \textit{think} part records the full reasoning trace, often including multiple thoughts after a series of trial and error and reflection. 
The \textit{answer} part is usually a concise summary based on the preceding \textit{think}, \textbf{containing only the main reasoning thought utilized in the \textit{think} that leads to the final answer.}
Assuming consistency between \textit{answer} and \textit{think}, we argue that improving reasoning quality primarily hinges on refining the \textit{think}. Therefore, our process evaluation focuses on the \textit{think} part instead of the \textit{answer} one.
For convenience, we will abuse the term \textit{solution} to represent \textit{think} in the following sections unless otherwise specified.

\section{Our Method}\label{sec:method}

In this section, we propose a new RL algorithm driven by process rewards. 
We first introduce a novel evaluating mechanism for intermediate reasoning steps in math problem solving \S(3.1). 
Based on this evaluation mechanism, we further develop an RL algorithm that effectively leverages these process rewards \S(3.2).
Finally, we present an efficient off-policy training framework \S(3.3). An overview of the entire method is illustrated in Figure 1.

\subsection{Intrinsic Signal-driven Thought-level Generative Process Evaluation}\label{sec:process_eval}

Before performing process evaluation, we decompose each solution into a sequence of steps with suitable granularity. While we prompt the LRM to perform such decompositions, alternative approaches are equally applicable, as our ultimate goal is to perform evaluation at the thought level.

\begin{figure*}[t]
    \centering
    \includegraphics[width=\linewidth]{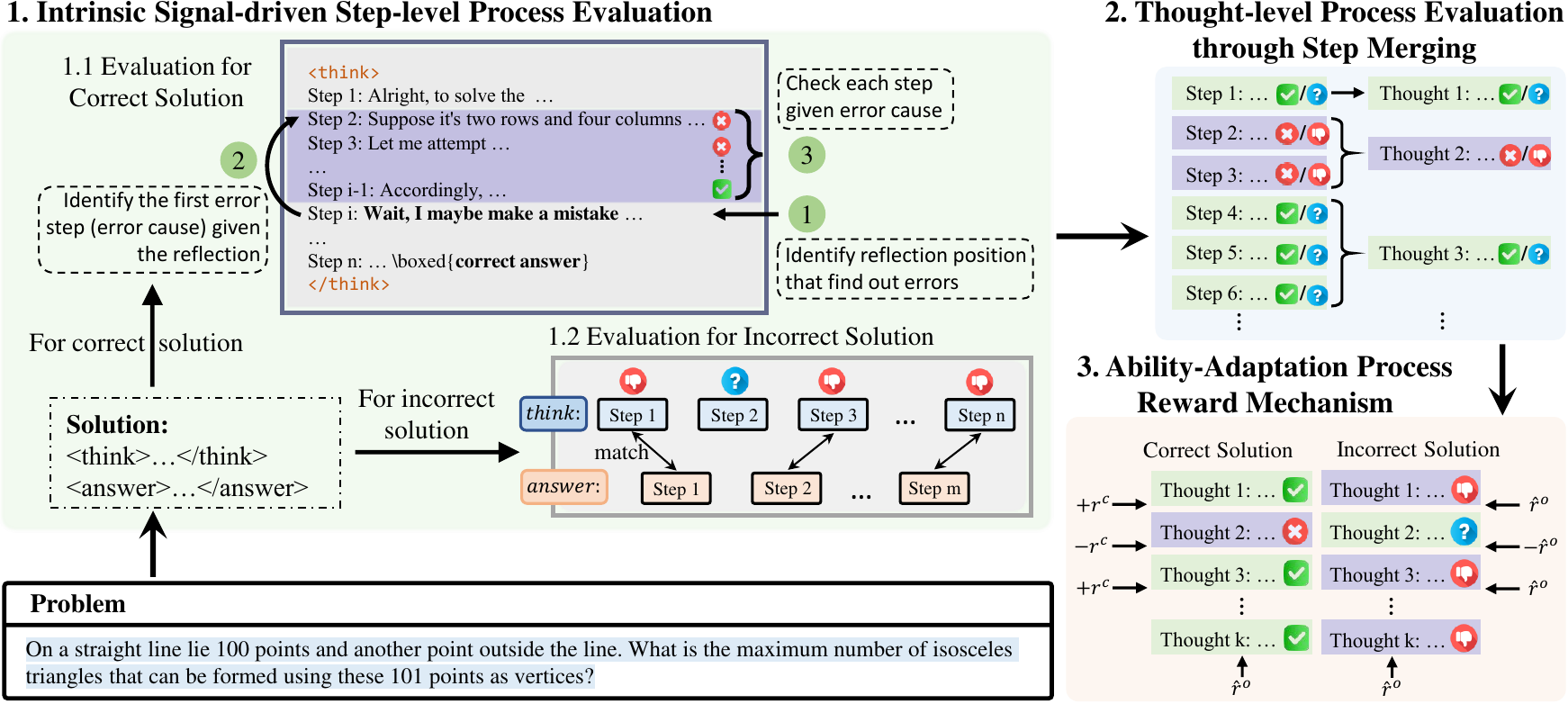}
    \caption{
    The pipeline of intrinsic signal-driven thought-level generative process evaluation. The whole process consists of three steps: (1) Intrinsic Signal-driven Step-level Process Evaluation, which evaluates each reasoning step using intrinsic signals from solutions; (2) Thought-level Process Evaluation through Step Merging, which merges continuous steps with the same correctness to enable thought-level evaluation; and (3) Ability-Adaptation Process Reward Mechanism, which calculates rewards based on model capability, balancing exploration and exploitation.
    }
    \label{fig:process_evaluation}
% \vspace{-0.3cm}
    % \vskip-10pt
\end{figure*}

Then, we propose a systematic three-stage framework for evaluating reasoning processes. The 1st stage introduces a tailored prompting pipeline to assess the correctness of each reasoning step, with reduced reliance on the evaluator’s reasoning capabilities. The 2nd stage aggregates step-level evaluations into thought-level assessments, effectively mitigating challenges arising from ambiguous step segmentation and noisy judgments. Finally, the 3rd stage incorporates a capacity-adaptive reward mechanism that assigns fine-grained rewards based on the correctness of each reasoning thought. An overview of the evaluation pipeline is provided in Figure~\ref{fig:process_evaluation}.

\subsubsection{Stage 1: Intrinsic Signal-driven Step-level Process Evaluation}\label{sec:step-level_process_eval}
% Traditional process evaluation methods typically decompose solutions into steps and apply a Process Reward Model to assess the correctness of each step independently. 
% In contrast, we propose two distinct evaluation procedures based on the correctness of solutions, reflecting our different reward expectations.
% Similar to existing generative PRMs, we also begin with step-level evaluation. Prior research includes training dedicated generative PRMs or prompting PRMs to directly assess the correctness of each step sequentially. 
% However, these methods potentially assume that the LRM-based PRM must, in principle, possess the ability to solve the original problem, thereby raising the bar for PRM design.
% In contrast, we propose a new method driven by the reasoning capabilities of LRMs. To lower the complexity of the process evaluation task, we decompose or reformulate it from various perspectives into simpler reasoning subtasks, allowing the LRM’s built-in reasoning skills to be more effectively applied for process evaluation. 
% In this work, we propose a novel evaluation method driven by the reasoning capabilities of the LRM. 
% To minimize the demands on the PRM, we decompose or reformulate the original evaluation task from different perspectives into simpler reasoning subtasks.
% Based on the correctness of solutions, we propose two distinct evaluation procedures, reflecting different expectations for process rewards.

Similar to existing generative PRMs, our approach starts with step-level evaluation. Prior work has explored training specialized generative PRMs or prompting LRM to directly assess the correctness of each reasoning step. However, these methods implicitly assume that the LRM-based PRM must possess the reasoning capability to solve the problem—a requirement that significantly increases the difficulty of process evaluation.
% \mrc{We instead propose a novel evaluation paradigm that relies on comprehension rather than reasoning capacity. 
% Based on the correctness of solutions, we design two distinct evaluation procedures to reflect different expectations for process rewards: for correct solutions, we identify incorrect reasoning thoughts; for incorrect ones, we match \textit{think} with the \textit{answer} to locate the key thought responsible for the error.}
We introduce a new evaluation paradigm based on intrinsic signals in Long CoT. The PRM assesses reasoning steps mainly through comprehension rather than reasoning, drastically lowering the capability requirements for PRM.
Based on the correctness of solutions, we design two distinct evaluation procedures to reflect different expectations for process rewards: 

\textbf{Case 1: Correct solutions.}
% During training, if the current LRM is already able to solve a problem, LRM is required to be able to summarize the correct solution into more concise knowledge, which can serve as a reasoning foundation knowledge when tackling more challenging problems. Inspired by how students reflect on correct solutions to identify which parts were useful or erroneous, we encourage the LRM to selectively learn from effective sub-trajectories.
% During training, once the LRM has sufficiently explored to solve the tasks, relying simply on final outcome rewards for optimization inevitably results in inefficient utilization of valuable solutions. Inspired by how students summarize knowledge after solving the questions by reviewing critical reasoning steps and disregarding incorrect attempts, we encourage the LRM to selectively learn from its solutions. 
We first aim to distinguish which steps within a correct solution are valid and which correspond to erroneous trials. 
% A straightforward approach is to prompt the LRM to sequentially assess the correctness of each step after thinking.
% To this end, we propose a 3-step prompt-based evaluation pipeline:
% However, we observe that for long solutions, the number of decomposed steps can be prohibitively large, rendering step-by-step evaluation inefficient. 
To address this, we decompose the evaluation into three simpler subtasks, proposing a 3-step evaluation pipeline.
\begin{enumerate}[leftmargin=1cm, label=\arabic*)]
    \item Prompt the PRM to identify the specific step where an error is recognized through reflection. This is relatively easy as the solution often includes intrinsic reflective triggers such as ``wait'' or ``I made a mistake'';
    \item Prompt the PRM to identify the earliest step responsible for this error—referred to as the error cause. Since the reflection thought in 1) often includes an analysis of the error cause, pinpointing this step is generally not difficult;
    \item Prompt the PRM to evaluate the correctness of each intermediate step between the error cause and the reflection step. Given both the error span and its reason, evaluating the correctness of each step becomes more manageable.
\end{enumerate}

% Compared to directly evaluating each step in sequence, this localized reflective strategy improves evaluation efficiency by narrowing down the error region before detailed inspection.
% When evaluating the correctness of each reasoning step sequentially, we categorize steps into three types: correct, incorrect, and reflective or verificatory. 
% \mrc{Compared to directly evaluating each step using GenPRM, our strategy—first identifying the error range and then progressively refining the assessment—reduces the number of evaluation steps. 
% % This approach not only improves evaluation efficiency but also lowers the frequency of misevaluation in theory.
% We hope for this mechanism to reduce the frequency of misjudgments during training.}

This process naturally mirrors a learner’s review and reorganization of reasoning solutions after problem-solving. 
\textbf{Besides, to avoid suppressing cognitive behaviors such as reflection and self-correction during RL training, we treat the reflective steps as correct in Step 3).}

\textbf{Case 2: Incorrect solutions.}
% For incorrect solutions, the same evaluation strategy applied above is no longer applicable. This is because the reflective behaviors in incorrect solutions may be incorrect—otherwise, the LRM would have been able to reason correctly. To address this, we propose a matching-based evaluation approach. Specifically, we compare each step in the \textit{think} with the \textit{answer}, classifying each step in \textit{think} as either matched or unmatched. Our core assumption is that, while \textit{think} may contain multiple exploratory or tentative reasoning thoughts, the one ultimately applied in the \textit{answer} must be incorrect.
For incorrect solutions, the evaluation strategy for correct solutions is no longer applicable, as the reflective steps in such cases may be flawed—otherwise, the solution would be corrected. Instead, we propose a matching-based evaluation approach. Specifically, we compare each step in the \textit{think} with the \textit{answer} and identify matched \textit{think} steps. Our core assumption is that while the \textit{think} may contain various exploratory reasoning paths, the one ultimately applied in the \textit{answer} must be incorrect, as the answer is incorrect.

By performing fine-grained, step-level matching between \textit{think} and \textit{answer}, 
we can label the matched steps as flawed, while unmatched steps are marked as uncertain. This matching-based evaluation mechanism significantly reduces the reliance on the evaluator's reasoning capabilities—requiring only the ability to understand a given reasoning trace rather than solving the math problem directly. 
In fact, we do not attempt to assess the precise correctness of every individual step, as such fine-grained evaluation imposes high reasoning demands on PRMs. Conversely, treating all steps in an incorrect solution as invalid—typical in outcome-only reward approaches—risks excessive penalization. Therefore, we pursue a principled trade-off: enabling selective penalization (compared to outcome-only methods) while significantly reducing the reasoning capability required (compared to accurate evaluation for each step).
As a result, our method remains effective and generalizable even when high-capacity reasoning models are not available.
All detailed prompts used are provided in Appendix~\ref{app:prompt_details}.

\subsubsection{Step 2: Thought-level Process Evaluation through Step Merging}\label{sec:thought-level_process_eval}

By merging the step-level evaluation results, we further obtain thought-level process evaluation. The implementation is conceptually simple: for correct solutions, consecutive correct or incorrect reasoning steps are merged into correct or incorrect units, named as `\textbf{thoughts}', respectively; for incorrect solutions, consecutive steps that are matched or unmatched with the \textit{answer} are grouped into matched or unmatched thoughts. Instead of assigning per-step rewards, we assign a single reward at the end of each thought segment, as detailed in Stage 3.
Our thought-level evaluation offers the following advantages:

\begin{itemize}[leftmargin=1cm]
    \item \textbf{Clearer process segmentation:} 
    % \mrc{Unlike the often ambiguous step-level boundaries, thoughts are segmented based on correctness, resulting in more semantically coherent partitions.} 
    Thoughts are split based on correctness, yielding more coherent and interpretable segments compared to ambiguous step-level divisions.
    % Moreover, these boundaries frequently align with key discourse markers such as ``wait'' or ``alternatively'', making the segmentation more clearer.
    \item \textbf{Mitigating noisy reward signals.} 
    % Step-level rewards are prone to misjudgment due to the limited accuracy of process evaluators, especially when the number of steps is large. Aggregating steps into thoughts smooths such noise and reduces the impact of incorrect rewards.
    % The accuracy of PRM evaluations is limited, and the number of steps in Long CoT far exceeds the number of thoughts. Therefore, relying on step-level rewards amplifies evaluation noise. 
    % Aggregating at the thought level mitigates this issue by smoothing over noisy step-wise assessments.
    % Thought-level evaluation mitigates the impact of erroneous reward signals by introducing fewer opportunities for reward misassignment.
    % \mrc{The accuracy of PRM evaluations is inherently limited, and the number of steps in Long CoT often far exceeds the number of thoughts. As a result, assigning rewards at the thought level can substantially reduce the impact of evaluation noise.}
    Given the limited accuracy of PRMs and the typically large number of reasoning steps in Long CoT, evaluating at the thought level significantly reduces the impact of noisy evaluations.

    \item \textbf{Resistance to reward hacking:} A common reward hacking issue arises when the LRM obtains more reward by repeatedly generating positive patterns. 
    % By assigning rewards at the thought level—i.e., to merged step blocks—repetitive or redundant reasoning no longer accumulates additional reward, discouraging such shortcuts.
    % By combining consecutive steps of the same evaluation type, repeated generation of positively rated steps does not lead to additional rewards.
    Merging homogeneous steps prevents such behavior by withholding additional reward for repeated positive evaluations.

    \item \textbf{Structured segmentation for RL.} Through step merging, 
    % \mrc{correct solutions are segmented into alternating correct and incorrect thoughts, while incorrect solutions are divided into alternating matched and unmatched thoughts. This segmentation lays the groundwork for designing new RL algorithms that more effectively leverage process-level rewards, which is detailed in~\ref{sec:final_grpo_algo}.}
    solutions are segmented into alternating correct and incorrect thoughts or alternating matched and unmatched thoughts-forming a structured basis for developing RL algorithms that better exploit process-level feedback, as further detailed in section~\ref{sec:final_grpo_algo}.
\end{itemize}
% In the following, we describe how to assign concrete reward values for all thought-level evaluation results.

\subsubsection{Step 3: Ability-Adaptation Process Reward Mechanism}\label{sec:process_reward_calculate}
After performing process evaluation, 
% And the process reward is assigned only to the final token of each thought.
we further design two distinct process reward mechanisms based on the solution correctness.

\begin{figure*}[t]
    \centering
    \includegraphics[width=\linewidth]{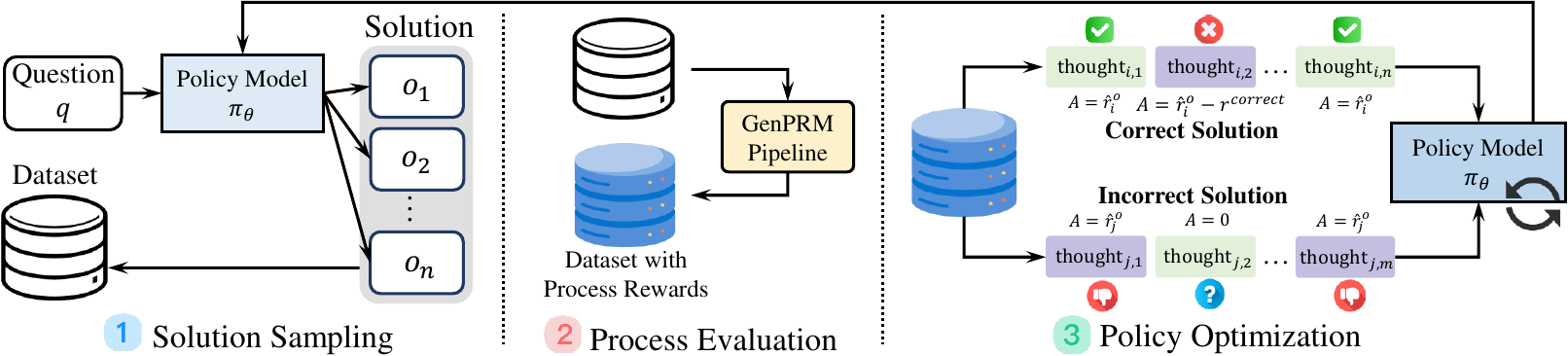}
    \caption{
    The overview of our proposed efficient off-policy training pipeline. This design ensures full utilization of GPU resources at each stage—all GPUs are either concurrently used for sampling, process evaluation, or parameter updates, avoiding any idle computation.
    }
    \label{fig:tp_grpo}
% \vspace{-0.3cm}
    % \vskip-10pt
\end{figure*}

% \begin{itemize}[leftmargin=1cm]
\paragraph{For Correct solutions.} 
% For a correct solution $o_i$, we encourage the LRM not only to maintain its correctness but also to avoid erroneous reasoning and to identify effective solutions more efficiently. 
% To this end, we assign the same positive reward $+r^{\text{correct}}$ to all correct thoughts and the same negative reward $-r^{\text{correct}}$ to all incorrect ones.
For a correct solution $o_i$, we assign the same positive reward $+r^{\text{correct}}$ to all correct thoughts and the same negative reward $-r^{\text{correct}}$ to all incorrect ones.

% The design of $r_{\text{correct}}$ balances exploration and efficiency. When a problem is difficult for the current LRM, we encourage diverse reasoning and reflection, favoring outcome-level rewards (i.e., process rewards should be small). When the task is easier, we emphasize efficiency by suppressing redundant or incorrect reasoning through larger process rewards. 
% The design of $r^\text{correct}$ needs to strike a balance between exploration and exploitation. When a problem is difficult for the current LRM, it may occasionally reason the correct solution within $G$ attempts, but the model has not yet mastered the task. In such cases, we still encourage diverse exploration, so the reward should primarily come from the final outcome and $r_\text{correct}$ should remain small. In contrast, once the LRM can reliably solve the problem, we expect it to refine and consolidate its reasoning. At this stage, we assign larger process rewards to penalize redundant or erroneous reasoning, thereby enhancing learning efficiency.
The design of $r^\text{correct}$ needs to strike a balance between exploration and exploitation. 
% When a problem is hard for the current LRM, we prioritize encouraging exploration, including more trial and error; thus, rewards should primarily stem from the outcome reward, and $r^\text{correct}$ should be kept small.
When a problem is hard for the current LRM, we make rewards primarily stem from the outcome by keeping $r^\text{correct}$ small, thereby prioritizing encouraging exploration, including more trial and  error.
In contrast, once the LRM can reliably solve the problem, the focus shifts toward refining and consolidating the reasoning process. And we assign larger process rewards to penalize erroneous reasoning, thereby enhancing learning efficiency.
To achieve the above targets, we define $r_\text{correct}$ as:
  \begin{equation}
      r^{\text{correct}} = \alpha \cdot \text{acc\_mean}
  \end{equation}
where $\alpha > 0$ is hyper-parameter to control the strength of process rewards, and $\text{acc\_mean}$ is the average accuracy of $G$ solutions within the same group sampled in GRPO~\cite{shao2024deepseekmath}. $\text{acc\_mean}$ reflects the LRM’s capability on the current problem. When all solutions are incorrect, $r^{\text{correct}} = 0$ and the setup reduces to the Outcome-only GRPO; when all solutions are correct, $r^{\text{correct}} = \alpha$ and the strongest process reward is applied to reinforce or suppress specific thoughts. We discuss the optimization implications of this design in Section~\ref{sec:final_grpo_algo}.

\paragraph{For Incorrect solutions.} For an incorrect solution $o_i$, we penalize only the reasoning thoughts in \textit{think} directly responsible for the error—i.e. thoughts matched with the \textit{answer}. Matched thoughts are penalized with $+r^{\text{incorrect}}$ at their final token, while unmatched thoughts receive $-r^{\text{incorrect}}$. We define incorrect reward $r^{\text{incorrect}}$ as:
\begin{equation}
      r^{\text{incorrect}}_i = \hat{r}_i^o
\end{equation}
where $\hat{r}_i^o$ is the normalized outcome reward of solution $o_i$ defined in Eq.~\ref{eq:reward_normalize}. 
% This ensures that thoughts unmatched to the \textit{answer} receive zero advantage and are not penalized, avoiding suppression of potentially useful, unexplored reasoning. 
This ensures that thoughts unmatched to \textit{answer} are not penalized, thereby avoiding the suppression of potentially useful but not fully explored reasoning thoughts~\cite{wang2025thoughts}.
We further explain this reward design in Section~\ref{sec:final_grpo_algo}.
% \end{itemize}

\subsubsection{Limitations}
The preceding section outlines our proposed process evaluation framework. While it introduces several novel mechanisms, certain limitations remain:
\begin{itemize}[leftmargin=1cm]
    \item The primary limitation lies in the 1st-stage step-level evaluation. Due to its inherently multi-step procedure, this stage exhibits considerable instability—an issue commonly observed in GenPRM approaches that lack dedicated task-specific training. To address this, future work will explore the design of more robust and instruction-sensitive prompts, as well as the fine-tuning of GenPRMs to enhance evaluation reliability and consistency.
    \item The proposed process evaluation method requires multiple calling for the GenPRM, resulting in low evaluation efficiency. This is also a common limitation for previous GenPRMs. To mitigate this issue, we introduce an off-policy training pipeline introduced in the following that minimizes GPU idle by decoupling sampling and optimization. Additionally, we employ heuristic rules to filter solutions committed to process evaluation—such as those with excessively short \textit{think} or \textit{answer} content—thereby improving overall efficiency.
    \item 
    Moreover, for incorrect solutions, our approach does not exactly verify the correctness of each intermediate step. Instead, it penalizes all tokens within the primary thought deemed responsible for the final error. It inevitably results in misattribution of penalties to partially correct reasoning steps in the primary thought, as not all steps within the primary thought are erroneous. We regard this as a pragmatic compromise between two extremes: coarse-grained outcome-level rewards, which indiscriminately penalize all tokens, and fine-grained step-level rewards, which require the high capacity of precise error localization. This trade-off balances the evaluation accuracy with the practical constraints on the PRM’s capacity.
    % We hypothesize that, as RL scales, such a strategy may progressively approximate the error sources—similar to outcome-based RLVR—but with the potential for faster convergence due to the availability of intermediate process-level signals.
\end{itemize}

\subsection{TP-GRPO: GRPO based on Thought-level Process Reward}\label{sec:final_grpo_algo}
% We train the LRM using the process-reward-augmented GRPO algorithm mentioned in Section~\ref{sec:grpo_foundation}. 
Finally, we integrate the proposed process evaluation framework with GRPO~\citep{guo2025deepseek}—the most widely adopted RL algorithm in current LLM training—and introduce a new algorithm, TP-GRPO. TP-GRPO jointly optimizes the LLM under the guidance of both process and outcome rewards.
For outcome rewards, we retain the standard group-level normalization as in Eq.~\ref{eq:reward_normalize}.
% \begin{equation}
%     \hat{r}_{i}^o = \frac{r_i^o - \text{mean}(\{r_j^o\}_{j=1}^G)}{\text{std}(\{r_j^o\}_{j=1}^G)}
% \end{equation}
For correct solutions $o_i$, we assign $+r^{\text{correct}}$ to the final token for each correct thought and $-r^{\text{correct}}$ to that of each incorrect thought.
For incorrect solutions, we assign $+r^{\text{incorrect}}_i$ to thoughts matched with the answer, and $-r^{\text{incorrect}}_i$ to unmatched thought segments. 
The final thought segment receives no process reward, as its last token already receives the outcome reward. 
In summary, we formalize the reward assignment as follows:
\begin{equation}
    r_{i,t}=
    \begin{cases}
    \hat{r}_{i}^o, & o_{i,t} \text{ is the final token of solution } o_i \\
    +r^{correct}, & o_i \text{ is correct } \wedge o_{i,t} \text{ is the end token of a correct thought} \\
    -r^{correct}, & o_i \text{ is correct } \wedge o_{i,t} \text{ is the end token of an incorrect thought} \\
    +r_i^{incorrect}, & o_i \text{ is incorrect } \wedge o_{i,t} \text{ is the end token of a matched thought} \\
    -r_i^{incorrect}, & o_i \text{ is incorrect } \wedge o_{i,t} \text{ is the end token of an unmatched thought} \\
    0, & \text{else} 
    \end{cases}
\end{equation}
Then, the advantage of each token $o_{i,t}$ is calculated as $A_{i,t}=\sum_{t'=t}^{|o_i|}r_{i,t'}$ following the process supervised GRPO~\citep{shao2024deepseekmath}, where $|o_i|$ is the number of tokens in $o_i$.
Finally, the optimization objective is defined in Eq.~\ref{eq:grpo_loss}:
% Finally, the optimization objective is defined as process supervised GRPO~\cite{shao2024deepseekmath}:
% \begin{equation}
%     \begin{aligned}
%         \mathcal{J}_{\text{GRPO}}(\theta) &= \mathbb{E}[{q\sim P(Q), \{o_i\}_{i=1}^G\sim\pi_{\theta_{old}}(Q|q)}]\\
%         &\frac{1}{\sum_{i=1}^G|o_i|}\sum_{i=1}^G\sum_{t=1}^{|o_i|} \{ \min[\rho_{i,t} \hat{A}_{i,t},
%         \text{clip}(\rho_{i,t}, \\
%         &1-\epsilon_{\text{low}}, 1+\epsilon_{\text{high}})\hat{A}_{i,t}] - \beta D_{\mathrm{KL}}[\pi_\theta \| \pi_{\mathrm{ref}}]\},
%     \end{aligned}\label{eq:grpo_loss}
% \end{equation}
% where $\epsilon_{\text{high}}$ and the token-level loss are two modifications proposed in DAPO~\cite{yu2025dapo}.

To better understand the design of these two types of process rewards, we establish the following theoretical properties from the perspective of advantage estimation in GRPO:

\begin{proposition} 
For a correct solution $o_i$, token $o_{i,t}$ in correct thoughts have advantage $A_{i,t}=\hat{r}_i^o$; while token $o_{i,t}$ in incorrect thoughts have advantage $A_{i,t}=\hat{r}_i^o - r^{\text{correct}}$.
\end{proposition}

\textit{Implication:} 
% This proposition shows that Correct thoughts retain the same advantage as under outcome-only reward, ensuring that the original optimization objective—producing correct final answers—is preserved. In contrast, the advantage values for incorrect thoughts are reduced compared to using outcome rewards alone, which implicitly discourages the tokens associated with flawed reasoning.
This proposition shows that tokens within correct thoughts retain the same advantage as using outcome-only rewards, preserving their original optimization objective during process-reward-augmented RL training. In contrast, tokens within incorrect thoughts receive reduced advantage values compared with using outcome-only rewards, effectively attenuating the optimization of flawed reasoning.

\begin{proposition}
For an incorrect solution $o_i$, token $c_t$ in matched thoughts have advantage $A_{i,t}=\hat{r}_i^o$, while token $c_t$ in unmatched thoughts have advantage $A_{i,t}=0$.
\end{proposition}

\textit{Implication:} 
% This result indicates that the advantage values of tokens in aligned thoughts are identical to those under outcome-only rewards, thereby preserving the original optimization objective. In contrast, uncertain thoughts receive zero advantage, avoiding the risk of suppressing potentially useful reasoning.
This proposition demonstrates that tokens within matched thoughts retain the same advantage values as under outcome-only rewards, thereby achieving the objective also mentioned above: introducing process rewards without altering the original training objective for these incorrect tokens during RL training. In contrast, unmatched thoughts receive 0 advantage, avoiding unintended penalization since their correctness remains uncertain.

The proofs of these two propositions are presented in Appendix~\ref{app:proof}.

\noindent \textbf{The importance of preserving the original optimization objective.} 
% \mrc{A well-recognized risk of incorporating process rewards is reward hacking, which we attribute to the shift of advantage estimation using process rewards. In math reasoning, the ultimate goal is to derive the correct answer. However, introducing process rewards can distort this objective, such as repeatedly generating steps that yield positive process rewards without contributing to correct reasoning.
% Our reward design explicitly avoids this pitfall. It suppresses incorrect thoughts in correct solutions and avoids optimizing uncertain thoughts in incorrect ones. At the same time, it preserves the intended learning target: promoting correct thoughts in correct solutions and discouraging incorrect thoughts in incorrect solutions, while minimizing unintended optimization bias.}
A widely discussed challenge in integrating process rewards is reward hacking, which we attribute to distortions in the advantage estimation induced by process rewards.While the ultimate goal in math reasoning is to arrive at the correct final answer, process rewards may inadvertently distort the LRM’s optimization objective.
Specifically, LRM may exploit the process reward to inflate the overall advantage estimate, leading the optimization objective away from deriving the correct answer—manifesting, for example, in the repeated generation of steps that yield high process rewards without making meaningful progress toward the solution.

Our reward design explicitly avoids this pitfall. Taking a correct solution as an example, TP-GRPO suppresses error thoughts without altering the advantage of tokens within the correct thoughts. As a result, the optimization objective for these tokens in key thoughts (correct or matched) remains aligned with that under the outcome reward alone—namely, improving reasoning accuracy—thus preventing any distortion from the ultimate learning objective.
% Finally, we apply the process and outcome reward in the GRPO training framework as described in Section~\ref{sec:preliminary}.

\subsection{Efficient Off-Policy Training Pipeline}
% Leveraging LLMs for process-level evaluation introduces significant efficiency challenges. In particular, an on-policy training approach leads to substantial GPU waiting time due to the significant slow process evaluation phase. To address this, we design an off-policy training pipeline that improves utilization and throughput. Each training iteration consists of three stages:
Prompting LLMs for process evaluation introduces significant efficiency challenges, as it requires multiple inference calls to the GenPRM. This inevitably leads to substantial GPU idle time under an on-policy training paradigm. To address this issue, we design an off-policy training pipeline that improves GPU utilization and accelerates training. This pipeline consists of three stages within each iteration:

\begin{enumerate}[leftmargin=0.5cm]
    \item \textbf{Stage 1: Solution Sampling.} We deploy the latest updated actor LRM using all GPUs to sample candidate solutions for each problem. Suppose Stage 3 performs $S$ training steps, each with $M$ problems, and each problem generates $G$ candidate solutions. Then, Stage 1 produces a total of $S \times M \times G$ solutions for $S \times M$ problems.
    \item \textbf{Stage 2: Process Evaluation.} We deploy multiple LLMs using all GPUs as the evaluator and evaluate the solutions using a multi-process, multi-threaded setup. This stage computes process rewards for $S \times M \times G$ sampled solutions.
    \item \textbf{Stage 3: Policy Optimization.} We train the actor LRM via the RL algorithm using all GPUs, based on the solutions and their corresponding process and outcome rewards from Stage 1 and 2. We filter out solutions without annotated process rewards after Stage 2 to focus on process rewards.
    % Since our primary focus is on studying the impact of process rewards, we filter out solutions without annotated process rewards after Stage 2.

\end{enumerate}
% After Stage 3, the newly updated actor model is used in the next Stage 1 for solution sampling.

\section{Experiments}
In this section, we evaluate the effectiveness of TP-GRRO in improving LRMs' math reasoning. 
% Beyond overall performance, we analyze the quality of our proposed process evaluation method and examine the influence of different evaluation backbones and reward configurations. 
Beyond accuracy assessment, we further examine training efficiency to validate that our process reward formulation improves the exploitation of reasoning trajectories.
% Due to resource constraints, experiments are conducted primarily on DeepSeek-R1-Distill-Qwen-1.5B and 7B~\citep{guo2025deepseek}. 
Given computational limitations, we focus our experiments on smaller LRMs. 
To meet the format requirement that explicitly includes both \textit{think} and \textit{answer} components, we select the DeepSeek-R1-Distill-Qwen 1.5B and 7B models for experiments.
We leave broader studies across more scales and variants to future work.

\subsection{Experimental Settings}
\textbf{Baselines}\quad
We primarily compare TP-GRPO with the reproduction of GRPO using only outcome rewards, aiming to demonstrate the advantage of our proposed process rewards in terms of both performance and efficiency. Specifically, we evaluate whether TP-GRRO can exceed the performance of outcome-only RL methods while requiring fewer training prompts, thereby validating its efficiency.
During TP-GRRO training, to minimize the influence of outcome rewards, we exclude solutions that do not obtain any intermediate rewards after process evaluation. 
% This typically occurs when all reasoning steps in a solution are either entirely correct or entirely incorrect, resulting in only a single merged thought and thus only an outcome reward. 
Besides, we also include results from several open-source baselines trained on DeepSeek-R1-Distill-Qwen models as reference and also compare efficiency with them.

\noindent\textbf{Training Data and Evaluation Benchmark}\quad
We conduct all experiments on DeepScaler-40K~\cite{deepscaler2025}. 
% We control the number of training prompts via the number of training steps. 
Due to limitations in computational resources, we do not train with full dataset. 
% Instead, we control the number of training prompts by setting the number of training steps. 
% We aim to compare training efficiency by measuring steps. 
% To maintain experimental consistency, all runs were conducted with an identical random seed 42.
We follow prior work and evaluate on 5 widely-used math benchmarks: AIME 24~\citep{AIME24}, AIME 25~\citep{AIME25}, AMC 23~\citep{li2024numinamath}, MATH-500~\citep{hendrycks2021measuring}, and Olympiad~\citep{he2024olympiadbench}.
For experiments with DeepSeek-R1-Distill-1.5B, we perform 16 independent Pass@1 evaluations and report the average (Avg@16) as final results. Due to the higher inference cost of DeepSeek-R1-Distill-7B and limited sizes of AIME 24, AIME 25, and AMC 23, we report Avg@5 on these subsets and Pass@1 on MATH-500 and Olympiad.
Besides, we aim to further investigate improvements in training efficiency. Due to variations in RL frameworks (e.g., TRL~\citep{vonwerra2022trl} and VeRL~\citep{sheng2025hybridflow}) and training settings, different methods are not directly comparable in terms of step number. 
% To ensure a fairer comparison, we report the number of training prompts (\#TP) processed.
To ensure a fairer comparison, we design a new metrics \textbf{\textit{Effic.}} to measure the training efficiency: $\textbf{\textit{Effic.}}=\frac{\text{Improvement}}{\#\text{training solutions}}*10^5$.

\noindent\textbf{Hyperparameters}\quad
All training experiments are conducted using the TRL framework and vLLM backend. We use a batch size of 5, a fixed learning rate of $1e-6$, and sample 8 rollouts per prompt. For TP-GRPO, we set the KL loss coefficient $\beta = 0$ following previous practice. 
The reproduced GRPO also uses the token-level loss and $\epsilon_{\text{high}}=0.28$ setting introduced in DAPO~\citep{yu2025dapo}.
For on-policy experiments, we save checkpoints every 50 training steps and perform evaluations accordingly. For off-policy experiments, we first collect solutions for 50 training steps at each iteration, then perform unified solution sampling and process evaluation. 
% As noted earlier, TP-GRRO filters out solutions without process rewards during training, so the actual number of training steps per iteration is typically lower than 50. 
More training and evaluation settings can be found in the Appendix~\ref{app:experiment_settings}.
% During evaluation, we use fixed settings with temperature$=0.6$, top-p$=0.95$, and maximum length$=32768$.

\subsection{Main Results}
\begin{table}[t]
    \centering
    \resizebox{\linewidth}{!}{
        \begin{tabular}{l| c c c c c| c}
            \toprule
            \multirow{2}{*}{\textbf{Model}} & \textbf{AIME 24} & \textbf{AIME 25} & \textbf{AMC 23} & \textbf{MATH-500} & \textbf{Olympiad} & \multirow{2}{*}{\textbf{Avg.}} \\
            & Avg@16 & Avg@16 & Avg@16 & Avg@16 & Avg@16 & \\
            % \midrule
            % \midrule
            % \multicolumn{8}{c}{1.5B Model} \\
            \midrule
            Base Model & 28.80 & 22.50 & 62.90 & 82.80 & 43.30 & 48.06 \\
            \midrule
            % \rowcolor{lightgray} 
            \multicolumn{7}{c}{Outcome Reward-based RLVR methods} \\
            \midrule
            % L1 & & & & & & & \\
            AutoThink (Stage 1) & 28.13 & 19.58 & 60.39 & 79.74 & 40.44 & 45.66 \\
            Open-RS1 & 30.42 & 20.83 & 63.78 & 83.01 & 43.91 & 48.39 \\
            Still-3-1.5B-Previous & 32.50 & 23.33 & 66.94 & 84.19 & 45.96 & 50.58 \\
            AdaptThink & 33.75 & 24.38 & 69.35 & 84.18 & 45.04 & 51.34 \\
            % FastCuRL-1.5B-Preview & 43.10 & 26.66 & 74.20  & 88.00 & 50.40 & 56.47 \\
            AReaL-1.5B-Preview(Stage 1) & 30.83 & 23.75 & 66.64 & 84.33 & 45.51 & 50.21 \\
            DeepScaler-1.5B-Preview & 43.10 & 29.38 & 73.60 & 87.80 & 50.00 & 56.78 \\
            \midrule
            \midrule
            {GRPO Replication (850 steps)} & 32.71 & 24.58 & 64.53 & 82.70 & 43.67 & 49.64 \\
            % {Off-policy + GRPO} & & & & & & \\
            \midrule
            \multicolumn{7}{c}{Process Reward-based methods} \\
            \midrule
            TP-GRPO (140 steps) & 33.12 & 25.63 & 64.01 & 83.81 & 43.91 & 50.10 \\
            % \midrule
            \bottomrule
        \end{tabular}
    }
    \vspace{8pt}
    \caption{Overall performance based on DeepSeek-R1-Distill-Qwen-1.5B.}
    \label{tab:main_results_1}
\end{table}
\begin{table}[t]
    \centering
    \renewcommand*{\arraystretch}{1.1}
    \resizebox{\linewidth}{!}{
        \begin{tabular}{l| c c c c c| c}
            \toprule
            \multirow{2}{*}{\textbf{Model}} & \textbf{AIME 24} & \textbf{AIME 25} & \textbf{AMC 23} & \textbf{MATH-500} & \textbf{Olympiad} & \multirow{2}{*}{\textbf{Avg.}} \\
            & Avg@5 & Avg@5 & Avg@5 & Pass@1 & Pass@1 & \\
            \midrule
            % \midrule
            Base Model & 54.00 & 36.67 & 81.20 & 92.00 & 55.11 & 63.80 \\
            \midrule
            \multicolumn{7}{c}{Outcome Reward-based RLVR methods} \\
            \midrule
            % Light-R1-7B-DS & & & & & & \\
            AutoThink (Stage 1) & 52.67 & 32.67 & 77.59 & 90.00 & 54.07 & 61.40 \\
            ReasonFlux-F1 & 54.67 & 34.00 & 80.00 & 91.60 & 55.26 & 63.11 \\
            AdaptThink & 52.67 & 38.00 & 82.17 & 90.80 & 56.89 & 64.11 \\
            AReaL-boba-RL-7B & 60.00 & 44.67 & 86.02 & 93.60 & 60.89 & 69.04 \\
            Skywork-OR1-7B & 69.33 & 51.33 & 87.71 & 93.40 & 61.33 & 72.62 \\
            \midrule
            \midrule
            {GRPO Replication (400 steps)} & 58.67 & 38.00 & 81.20 & 92.40 & 56.44 & 65.34 \\
            % {Off-policy + GRPO} & & & & & & \\
            \midrule
            \multicolumn{7}{c}{Process Reward-based methods} \\
            \midrule
            TP-GRPO (214 steps) & 60.67 & 43.33 & 83.13 & 92.00 & 57.04 & 67.23 \\
            \bottomrule
        \end{tabular}
    }
    \vspace{8pt}
    \caption{Overall performance based on DeepSeek-R1-Distill-Qwen-7B.}
    % \vspace{-0.3cm}
    \label{tab:main_results_2}
\end{table}

\begin{figure}[t]
    \centering
    \includegraphics[width=0.8\linewidth]{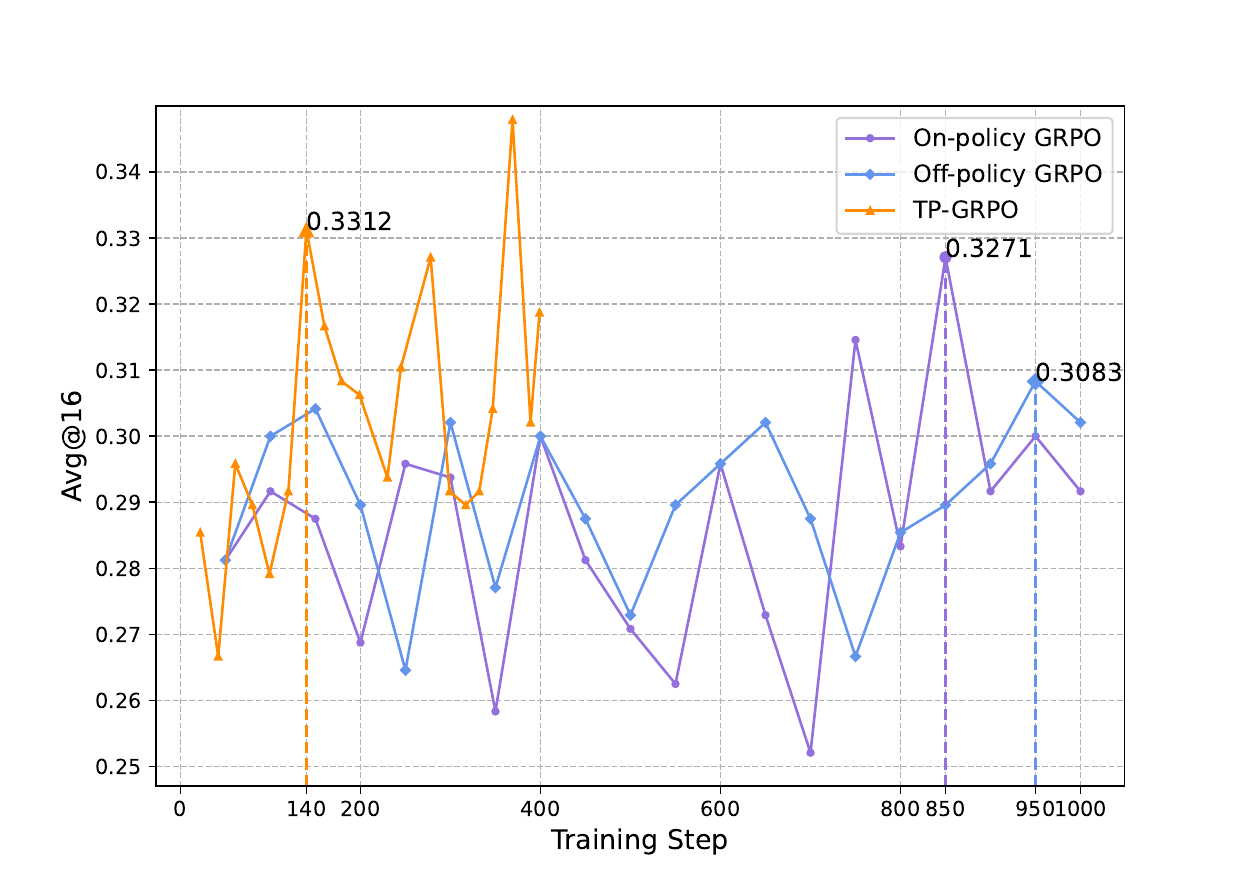}
    \caption{
    The trend of accuracy (Avg@16) on AIME 2024 over training steps for DeepSeek-R1-Distill-Qwen-1.5B base model.
    }
    \label{fig:acc_trend}
    % \vspace{-0.5cm}
    % \vskip-10pt
\end{figure}

Tables~\ref{tab:main_results_1} and~\ref{tab:main_results_2} present the experimental results using DeepSeek-R1-Distill-Qwen-1.5B and DeepSeek-R1-Distill-Qwen-7B as base models, respectively. 
% First, TP-GRPO consistently improves upon the base models: the 1.5B model achieves an average gain of +2.04 after training on only 700 prompts (140 steps). And the 7B model improves by +3.43 with just 1,070 prompts (214 steps). This demonstrates the effectiveness and efficiency of TP-GRPO in optimizing LRMs's math reasoning capabilities.
% Furthermore, TP-GRPO outperforms our reproductive on-policy GRPO baseline, which uses only outcome rewards. A more intuitive comparison is shown in Figure 3, which plots the trend of reasoning accuracy on AIME 24 over training steps. Here, off-policy GRPO refers to the GRPO pipeline with off-policy training but without process rewards. All methods evaluate performance every 50 training steps—except TP-GRPO, which uses a finer evaluation schedule. This is because TP-GRPO trains only on solutions that receive process rewards after process evaluation, rather than all sampled trajectories (see Appendix for detailed justification).
% As shown in Figure 3, TP-GRPO exhibits a significantly steeper learning curve, indicating improved optimization efficiency. This suggests that TP-GRPO enhances the exploitation of reasoning trajectories through well-designed process rewards, and enables LRMs to rapidly enhance their mathematical reasoning capabilities through fine-grained learning.
First, TP-GRPO achieves consistent improvements over the base models: on the 1.5B model, it attains an average gain of +2.04 after training on only 700 prompts (140 steps), with particularly strong improvements of +4.32 and +3.13 on AIME 2024 and AIME 2025, respectively. On the 7B model, TP-GRPO achieves an average improvement of +3.43 using just 1070 prompts (214 steps), including gains of +6.67 and +6.66 on AIME 2024 and AIME 2025. These results validate the efficacy of our thought-level process evaluation mechanism in guiding effective policy updates with limited data.
Moreover, TP-GRPO outperforms our reproductive on-policy GRPO baseline, which relies solely on outcome rewards. On the 1.5B model, TP-GRPO trained on 700 prompts surpasses the performance of on-policy GRPO trained on 4250 prompts. Similarly, on the 7B model, TP-GRPO with 1,070 prompts significantly exceeds the performance of on-policy GRPO after its 400 training steps. This substantial gain in learning efficiency highlights the advantage of incorporating structured process feedback.

A more intuitive comparison is provided in Figure~\ref{fig:acc_trend}, which plots the trend of reasoning accuracy (Avg@16) on AIME 2024 over training steps. Here, off-policy GRPO denotes a GRPO variant using the off-policy training paradigm but without process rewards. All baseline methods are evaluated every 50 training steps. In contrast, TP-GRPO exhibits a denser checkpoint schedule. This arises because, although TP-GRPO samples enough solutions per iteration to support 50 update steps, only a subset receives non-zero process rewards after evaluation. To more precisely analyze the role of process rewards in training, we filter out those solutions lacking process rewards. As a result, the effective number of training samples available in the 3rd stage of each iteration is reduced, which in turn limits the number of update steps that can be performed and leads to more frequent evaluation intervals.
As shown in Figure~\ref{fig:acc_trend}, TP-GRPO exhibits a significantly steeper learning curve, indicating faster convergence and higher training efficiency. This provides strong evidence that TP-GRPO enhances the exploitation of reasoning trajectories through well-designed process rewards, enabling efficient learning to strengthen the LRM’s math reasoning capabilities.
It should be noted that, for TP-GRPO, although a higher peak performance is observed after 140 steps, we find that the corresponding checkpoint shows limited improvement on AIME 2025. Considering overall training efficiency and cross-dataset generalization, we select the checkpoint saved at step 140 as the final TP-GRPO result. 

\begin{table}[t]
    \centering
    % \renewcommand*{\arraystretch}{1.1}
    % \resizebox{\linewidth}{!}{
        \begin{tabular}{l| c c c c}
            \toprule
            \textbf{Model} & \textbf{Traing Steps} & {\textbf{Batch Size}} & \textbf{Rollout Size} & \textbf{\textit{Effic.}} \\
            % \midrule
            % \midrule
            % \multicolumn{8}{c}{1.5B Model} \\
            \midrule
            \multicolumn{5}{c}{DeepSeek-R1-Distill-Qwen-1.5B} \\
            \midrule
            AutoThink (Stage 1) & 220 & 128 & 16 & -0.53 \\
            % L1 & & & & & & & \\
            AdaptThink & 300 & 128 & 16 & 0.53 \\
            Open-RS1 & 100 & 96 & 6 & 0.57 \\
            Still-3-1.5B-Previous & 250 & 128 & 8 & 0.98 \\
            AReaL-1.5B-Preview(Stage 1) & 200 & 1024 & 8 & 0.13 \\
            % FASTCuRL-preview & 860 & 128,64,64,64 & 8,8,8,16 & 8k,16k,24k,16k \\
            DS-1.5B-Preview & 1750 & 128,128,128 & 8,16,16 & $<$0.49 \\
            \midrule
            {GRPO Replication} & 850 & 5 & 8 & 4.65 \\
            % {Off-policy + GRPO} & & & & \\
            \midrule
            TP-GRPO & 140 & 5 & 8 & 36.43 \\
            \midrule
            \midrule
            \multicolumn{5}{c}{DeepSeek-R1-Distill-Qwen-7B} \\
            \midrule
            AutoThink (Stage 1) & 220 & 128 & 16 & -0.53 \\
            % L1 & & & & & & & \\
            ReasonFlux-F1 & 3750 & 16 & 1 & -1.15 \\
            AdaptThink & 150 & 128 & 16 & 0.10 \\
            AReaL-boba-RL-7B & 250 & 512 & 16 & 0.37 \\
            Skywork-OR1  & 1320 & 256,160 & 16,32 & $<$0.26 \\
            \midrule
            {GRPO Replication} & 400 & 5 & 8 & 9.6 \\
            % {Off-policy + GRPO} & & & & \\
            \midrule
            TP-GRPO & 214 & 5 & 8 & 40.07 \\
            % \midrule
            \bottomrule
        \end{tabular}
    % }
    \vspace{8pt}
    \caption{Training hyperparameters for efficiency comparison.}
    % \vspace{-0.3cm}
    \label{tab:main_results}
\end{table}

\subsubsection{More Efficiency Comparison}
% We also compare TP-GRPO with recent strong open-source RLVR approaches in Tables~\ref{tab:main_results_1} and~\ref{tab:main_results_2}. While our method lags behind SOTA models in absolute performance, this gap primarily stems from the training data scale. To enable a fairer comparison, Table 3 summarizes the training configurations of all methods for computing the efficiency metric (\textbf{\textit{Effic.}}). Notably, TP-GRPO achieves the highest performance gain per solution, surpassing even the best open-source model in terms of sample efficiency. This further validates the advantage of TP-GRPO in improving training efficiency and trajectory utilization.
We further compare TP-GRPO with recent strong open-source RLVR methods (see Tables \ref{tab:main_results_1} and \ref{tab:main_results_2}). Although our method lags behind the SOTA models in terms of absolute performance, this gap is primarily attributed to differences in the scale of training data used. RLVR approaches achieve RL scaling through extensive and efficient exploration. 
To enable a fairer comparison, Table~\ref{tab:main_results} summarizes the training settings of all methods with respect to the efficiency metric \textbf{\textit{(Effic.)}}. Notably, TP-GRPO achieves the highest \textbf{\textit{Effic.}}, outperforming all open-source baselines in terms of learning efficiency. This result further demonstrates the advantage of TP-GRPO in improving training efficiency and exploitation for reasoning trajectories.

\begin{table}[!h]
    \centering
    % \resizebox{0.85\linewidth}{!}{
    \begin{tabular}{l|ccccc}
    \toprule
    Model & AIME 24 & AIME 25 & Training Steps \\
    \midrule
    Base Model & 28.80 & 22.50 & - \\
    TP-GRPO & 33.12 & 25.63 & 140 \\
    - w/o CS Reward & 30.00 & 25.00 & 58 \\
    - w/o IS Reward  & 32.01 & 22.29 & 90 \\
    \bottomrule
    \end{tabular}
    % }
    \vspace{8pt}
    \caption{Ablation study on process rewards from correct vs. incorrect solutions. `CS reward' means `reward in \textbf{C}orrect \textbf{S}olutions', while `IS' means `\textbf{I}ncorrect \textbf{S}olutions'.}
    \label{tab:ablation_study}
\end{table}

\subsection{Analysis}
\subsubsection{Ablation Study}
We design two distinct process evaluation mechanisms for correct and incorrect solutions. 
% To better understand their individual contributions to the reward mechanism, we conduct ablation studies under two settings: 
To better understand their individual contributions, we conduct ablation studies under two settings: 
one using only process rewards from correct solutions, and the other using only those from incorrect solutions. The corresponding results are presented in Table~\ref{tab:ablation_study}.
First, both variants use fewer training steps compared to the full TP-GRPO setup. The reason is that more solutions are filtered out. The results in Table~\ref{tab:ablation_study} show that both process rewards yield measurable performance gains. 
% \st{Notably, the process reward based on incorrect solutions achieves improvement on both datasets—even with less training data. This may be attributed to the lower capability demand of evaluating incorrect solutions, which leads to more reliable and higher-quality process rewards. An additional intriguing observation is that the two reward types exhibit complementary performance patterns on AIME 24 and AIME 25. We plan to investigate this further in future work.}
% We observed an interesting phenomenon in our experimental results: on AIME 24, using only correct solution rewards yielded better performance than using only incorrect solution rewards. However, on AIME 25, the opposite was true.
% We believe this relates to the model's problem-solving ability. Our base model performed better on AIME 24, meaning it correctly solved more problems and thus benefited more from learning through correct solutions. In contrast, on AIME 25, most of the solutions generated by the model were incorrect (with an accuracy of only 22.50\%). Therefore, the model had more to gain from learning from incorrect solutions.
% Our experimental results demonstrate that our proposed process reward mechanism enables the model to learn effectively from both correct and incorrect solutions. 
We also observe an interesting phenomenon in our experimental results: on AIME 24, using only correct-solution rewards yields better performance, whereas on AIME 25, the opposite holds. We attribute this to the model’s differential problem-solving capability. The base model performs better on AIME 24, enabling more effective learning from correct solutions. In contrast, on AIME 25—where the base model’s accuracy is only 22.50\%, making incorrect solutions-driven learning more valuable.
Our results show that the proposed process reward mechanism supports effective learning from both correct and incorrect reasoning paths. 
% We hypothesize that dynamically adjusting the ratio of correct to incorrect solutions based on problem difficulty could further improve TP-GRPO’s performance—an avenue we leave for future work.
% We believe that dynamically adjusting the proportion of correct and incorrect solutions used for training, based on problem difficulty, could further enhance TP-GRPO's performance. We've left this for future work.
% \input{Tables/ablation_study}

\subsubsection{Different LRMs for Process Evaluator}
% To verify that our evaluation scheme imposes low reasoning requirements on the PRM, we conduct experiments using LLMs of different scales as the GenPRM. 
% % The corresponding results are shown in Figure 4. 
% % \input{tab}
% In the main experiments above, we use Qwen3-32B as the default GenPRM and further select two weaker LLMs, i.e. Qwen3-4B and Gemma-3-12b-it~\ref{team2025gemma}, as GenPRM, with GPQA accuracies of 65.6, 55.9, and 40.9\footnote{ Data from: \url{https://www.datalearner.com/ai-models}}, respectively. Results are shown in Figure~4. First, despite the significant differences in reasoning ability among these models, all GenPRMs push consistent improvements for 1.5B LRM via TP-GRPO. This demonstrates that our method has low dependency on the GenPRM’s reasoning proficiency. Moreover, comparable gains across model scales (Qwen3 32B vs. 4B) and families (Qwen3 vs. Gemma-3) further confirm the robustness of our designed process evaluation mechanism. 
To verify that our evaluation scheme imposes low reasoning requirements on the PRM, we conduct an analysis study using different LLMs as the generative PRM. In our main experiments, we adopt Qwen3-32B as the default GenPRM. For analysis, we additionally select two weaker LLMs—Qwen3-4B and Gemma-3-12B-IT \citep{team2025gemma}—as alternative GenPRMs. These three models exhibit progressively weaker reasoning abilities, as evidenced by their accuracy on GPQA Diamond~\citep{rein2024gpqa}: 65.6, 55.9, and 40.9, respectively\footnote{Data from: \url{https://www.datalearner.com/ai-models}}. We then apply TP-GRPO using each of the three GenPRMs. The results are shown in Table~\ref{tab:evaluator_variant}.
Despite the significant differences in reasoning capabilities among these models, all GenPRMs consistently improve the performance of the 1.5B LRM through TP-GRPO. This indicates that our method is relatively insensitive to the reasoning ability of the GenPRM. Furthermore, the comparable improvements achieved across different model scales (Qwen3-32B vs. 4B) and families (Qwen3 vs. Gemma) further validate the robustness of our proposed process evaluation mechanism.

\begin{table}[!t]
    \centering
    % \resizebox{\linewidth}{!}{
    \begin{tabular}{l|cccccc}
    \toprule
    Model & AIME 24 & AIME 25 & AMC 23 & MATH-500 & Avg. & Step \\
    \midrule
     Base Model & 28.80 & 22.50 & 62.90 & 82.80 & 49.25 & - \\
     TP-GRPO &  &  &  &  &  & \\
     - Qwen3-32B & 33.12 & 25.63 & 64.01 & 83.81 & 51.64 & 140 \\
     - Qwen3-4B & 33.33 & 24.58 & 63.78 & 83.62 & 51.33 & 155 \\
     - Gemma-3-12b-it & 32.71 & 23.33 & 64.76 & 83.73 & 51.13 & 125 \\
    \bottomrule
    \end{tabular}
    % }
    \vspace{8pt}
    \caption{Robustness Analysis with Different LLMs as PRMs.}
    \label{tab:evaluator_variant}
\end{table}

\begin{table}[!h]
    \centering
    % \resizebox{0.95\linewidth}{!}{
    \begin{tabular}{l|ccccc}
    \toprule
    Model & AIME 24 & AIME 25 & AMC 23 & Training Steps \\
    \midrule
    On-policy GRPO & 32.71 & 24.58 & 64.53 & 850 \\
    Off-policy GRPO & 30.83 & 24.79 & 62.05 & 950 \\
    TP-GRPO (25 steps) & 33.54 & 23.96 & 64.53 & 207 \\
    TP-GRPO (50 steps) & 33.12 & 25.63 & 64.01 & 140 \\
    \bottomrule
    \end{tabular}
    % }
    \vspace{8pt}
    \caption{Analysis study on the impact of off-policy training.}
    \label{tab:off_policy_compare}
\end{table}
\subsubsection{Analysis on Off-policy Training Pipeline}
% Due to the low efficiency of the process evaluation procedure, we propose an off-policy training pipeline to reduce training time. 
In this work, we propose an off-policy training pipeline to reduce training time to reduce GPU idling.
We further compare this framework with the on-policy method in terms of both reasoning performance and training efficiency, as shown in Table~\ref{tab:off_policy_compare}.
First, it is evident that off-policy GRPO underperforms on-policy GRPO in both accuracy and training efficiency, indicating that on-policy training is more advantageous—consistent with conventional training experience. Figure~\ref{fig:acc_trend} presents a clearer comparison, where on-policy GRPO demonstrates a more pronounced improvement trend over training steps.
To further investigate the impact of update frequency in off-policy training, we reduce the update interval of TP-GRPO's off-policy pipeline from 50 steps to 25 steps (and to 1 step in an extreme case, which essentially corresponds to true on-policy updates). We observe that both AIME 2024 and AMC 2023 tasks show further performance improvements with shorter update intervals.
In summary, although off-policy training is known to be less effective than on-policy training, TP-GRPO—trained under the off-policy setting—still outperforms the on-policy GRPO baseline. This further validates the effectiveness of our proposed GenPRM mechanism.

% \subsubsection{Hyperparameters Analysis}
% In computing the process rewards via Equation (5), we introduce two hyperparameters, $\alpha$ and $\beta$, where $\alpha$ controls the reward magnitude and $\beta$ modulates its variation trend. To investigate their impact on model performance, we adjust both parameters and report the results in Figure 5.

\section{Related Works}

\subsection{Process Reward Models}
PRMs evaluate reasoning trajectories by assigning rewards to each evaluation unit—typically a reasoning step-complementing traditional outcome rewards~\citep{li2024process, setlur2024rewarding, zhang2025lessons, younsi2025accurate}. Depending on the evaluation paradigm, PRMs are broadly categorized into discriminative and generative approaches~\citep{zhao2025genprm, zhong2025comprehensive}.
Discriminative PRMs predict step-level correctness scores via supervised learning~\citep{Skyworko1, ImpGenPRM}, often relying on large-scale human annotations such as in PRM800K~\citep{lightman2023let}. Recent works explore automatic labeling through self-training or simulation~\citep{wang2023math, luo2024improve, wang2024multi, lu2024autocv, lu2024autopsv}. However, these models often suffer from limited generalization, reward hacking, and poor interpretability~\citep{guo2025deepseek}.
In contrast, Generative PRMs frame evaluation as a generation task~\citep{ankner2024critique, zhao2025genprm}, assessing step correctness via generation and using the likelihood of indicative tokens (e.g., ``correct'', ``yes'') as rewards~\citep{zhang2024generative, mahan2024generative}. Leveraging LLMs’ inherent reasoning abilities, these methods offer better interpretability and scalability~\citep{wang2023chatgpt}. Many adopt an LLM-as-a-Judge framework~\citep{zheng2023judging, gu2024survey}, prompting pre-trained models without further fine-tuning~\citep{zheng2024processbench}, while others enhance performance via additional training~\citep{wang2024self, khalifa2025process}.
Despite progress, key challenges remain, including ambiguous step segmentation and excessive reliance on the PRM’s reasoning capabilities~\citep{guo2025deepseek, feng2025prm}. 
% To address these, we propose a thought-level evaluation framework and introduce a comprehension-driven GenPRM paradigm.

\subsection{PRMs for Optimizing Large Reasoning Models}
Prior to the release of DeepSeek R1~\citep{guo2025deepseek}, process rewards have been widely studied as a means to enhance the reasoning capabilities of LLMs~\citep{zhang2024rest, wang2024openr}. Depending on their application stage, process rewards are used either to guide trajectory pruning during inference~\citep{uesato2022solving, kalra2025verdict, muennighoff2025s1} or to shape training objectives~\citep{zhang2024o1}. This work focuses on the latter.
Early methods such as Rest-MCTS$^*$~\citep{zhang2024rest} train PRMs to rank reasoning trajectories and select high-quality paths for learning. \citet{chen2024autoprm} estimate step-level rewards via question decomposition but ultimately aggregate them into outcome rewards, limiting the effectiveness of fine-grained supervision. 
Similarly, \citet{li2024pspo} also fail to fully exploit process-level signals in their training pipeline. While OpenR~\citep{wang2024openr} employs PRMs to guide policy iteration in LLM training, it remains restricted to the discriminative paradigm.
% To unlock the full potential of PRMs, several challenges must be addressed—most notably, ambiguous step segmentation, susceptibility to reward hacking, and the need for continual retraining. 
To unlock the full potential of PRMs, we propose a novel evaluation mechanism, including a comprehension-driven evaluation strategy, a thought-level assessment granularity, and a capacity-adaptive reward mechanism.

\section{Conclusion and Future Work}

In this work, we introduce TP-GRPO, a novel RL framework guided by generative process rewards, with the goal of enhancing training efficiency in math reasoning tasks. 
% TP-GRPO tackles core limitations of conventional process reward approaches, ill-defined step boundaries, high dependency on evaluator reasoning capability and vulnerability to reward hacking. 
% To address these issues, we propose a comprehension-driven, thought-level evaluation mechanism, alongside a capacity-adaptive reward formulation that dynamically balances exploration and exploitation.
TP-GRPO tackles core limitations of conventional process reward approaches by proposing an intrinsic signal-driven, thought-level evaluation mechanism alongside a capacity-adaptive reward formulation that dynamically balances exploration and exploitation.
% Empirical results demonstrate that TP-GRPO achieves superior performance with significantly fewer training steps compared to outcome-only baselines, highlighting the effectiveness of process-rewards-guided optimization. 
Empirical results show that TP-GRPO achieves superior improvements compared to the outcome-only RLVR baseline under the same training sample budget, indicating that our designed process evaluation mechanism effectively enhances the optimization efficiency of LRMs.
% As a promising direction for future work, we plan to investigate how to jointly leverage outcome and process rewards to fully realize their complementary strengths in guiding both global objective alignment and fine-grained reasoning supervision.

% As a promising avenue for future research, we intend to explore dynamic strategies for alternating between process and outcome rewards to simultaneously enhance exploration and exploitation. 

However, there are inherent risks in overemphasizing process-level rewards. Focusing solely on exploitation while neglecting exploration can lead LRMs to overfit prematurely, thereby impairing their generalization capabilities. Moreover, our experiments do not demonstrate RL scaling through extensive training for TP-GRPO.
Nonetheless, as a study centered on the exploration–exploitation trade-off in RL, we argue that a promising direction lies in a dynamic strategy that alternates between process and outcome rewards. Such an approach can simultaneously enhance exploration and exploitation. This is analogous to how students expand the boundaries of their knowledge and abilities by exercising new problems, then consolidate their knowledge through reflection and review. Through this iterative cycle, consistent and efficient improvement can be achieved.

\bibliographystyle{plainnat}
\bibliography{neurips}

\appendix

\section{Proof of Proposition}\label{app:proof}

% \begin{proposition} 
\textbf{Proposition 1.}
\textit{For a correct solution $o_i$, token $o_{i,t}$ in correct thoughts have advantage $A_{i,t}=\hat{r}_i^o$; while token $o_{i,t}$ in incorrect thoughts have advantage $A_{i,t}=\hat{r}_i^o - r^{\text{correct}}$.}
% \end{proposition}

\begin{proof}
The proof is straightforward. After merging steps of the same type, the original \textit{think} can be decomposed into an alternating sequence of correct and incorrect thoughts:
$$
\texttt{think} = \text{thought}_1^{\text{correct}} \circ \text{thought}_2^{\text{incorrect}} \circ \cdots \circ \text{thought}_{n-1}^{\text{incorrect}} \circ \text{thought}_n^{\text{correct}}.
$$
% \begin{multline}
% \texttt{think} = \text{thought}_1^{\text{correct}} \circ \text{thought}_2^{\text{incorrect}} \circ \cdots \\
% \circ \text{thought}_{n-1}^{\text{incorrect}} \circ \text{thought}_n^{\text{correct}}.
% \end{multline}

Accordingly, the thought-level reward sequence is:
$$
[r_{\text{index}(1)}, r_{\text{index}(2)}, \dots, r_{\text{index}(n)}] = [r^{\text{correct}}, -r^{\text{correct}}, \dots, -r^{\text{correct}}, \hat{r}^{o}],
$$
% \begin{equation}
% \begin{split}
% [r_{\text{index}(1)}, r_{\text{index}(2)}, \dots, r_{\text{index}(n)}] ={} \\
% [r^{\text{correct}}, -r^{\text{correct}}, \dots, -r^{\text{correct}}, \hat{r}^{o}],
% \end{split}
% \end{equation}

where $\text{index}(j)$ denotes the index of the last token in the $j$-th thought. The last thought $t_n^{\text{correct}}$ receives only the outcome reward $\hat{r}^{o}$.

For tokens in a correct thought $\text{thought}_j^{correct}$, the advantage value is computed as:
\begin{equation}
    A^{\text{correct}}_{\text{index}(j-1)<k\le \text{index}(j)}
= r_{\text{index}(j)} + r_{\text{index}(j+1)} + \cdots + r_{\text{index}(n)}.
\end{equation}
Given the alternating reward pattern, this simplifies to:
$$
\underbrace{r^{\text{correct}} + \cdots + r^{\text{correct}}}_{(n-j)/2 \text{ terms}} + \underbrace{(-r^{\text{correct}}) + \cdots + (-r^{\text{correct}})}_{(n-j)/2 \text{ terms}}+\hat{r}^{o} = \hat{r}^{o}.
$$

% \begin{multline}
% \underbrace{r^{\text{correct}} + \cdots + r^{\text{correct}}}_{(n-j)/2 \text{ terms}} \\
% + \underbrace{(-r^{\text{correct}}) + \cdots + (-r^{\text{correct}})}_{(n-j)/2 \text{ terms}} + \hat{r}^{o} = \hat{r}^{o}.
% \end{multline}

Similarly, for tokens in the incorrect thought $\text{thought}_i^{incorrect}$, the corresponding advantage is:
\begin{equation}
\begin{aligned}
    A_{\text{index}(j-1)<k\le \text{index}(j)}^\text{correct}&=r_{\text{index}(j)}+r_{\text{index}(j+1)}+\cdots+r_\text{index}(n)\\
&=\underbrace{r^\text{correct}+\cdots + r^\text{correct}}_{(n-j+1)/2-1 \text{ times}}+(\underbrace{-r^\text{correct}-\cdots-r^\text{correct}}_{(n-j+1)/2\text{ times}})+\hat{r}^{o}\\
&=\hat{r}^{o}-r^\text{correct}
\end{aligned}
\end{equation}

% \begin{equation}
% \begin{aligned}
% A_{\text{index}(j-1)<k\le \text{index}(j)}^\text{correct} &= r_{\text{index}(j)}+r_{\text{index}(j+1)}+\cdots+r_\text{index}(n) \\
% &= \underbrace{r^\text{correct}+\cdots + r^\text{correct}}_{(n-j+1)/2-1 \text{ times}} \\
% & \qquad + (\underbrace{-r^\text{correct}-\cdots-r^\text{correct}}_{(n-j+1)/2\text{ times}}) + \hat{r}^{o} \\
% &= \hat{r}^{o}-r^\text{correct}
% \end{aligned}
% \end{equation}

\end{proof}

% \begin{proposition} 
\textbf{Proposition 2.}
\textit{For an incorrect solution $o_i$, token $c_t$ in matched thoughts have advantage $A_{i,t}=\hat{r}_i^o$, while token $c_t$ in unmatched thoughts have advantage $A_{i,t}=0$.}
% \end{proposition}

\begin{proof}
    Following the same reasoning as in Proposition 1, for an incorrect \textit{think}, we first merge steps of the same type and decompose the sequence into alternating matched and unmatched thoughts with respect to the \textit{answer}. We assign a penalty to matched thoughts with reward $r^{\text{incorrect}} = \hat{r}^o$, and reward unmatched thoughts with $-r^{\text{incorrect}} = -\hat{r}^o$.

For a matched thought $t_j^{\text{matched}}$, the advantage of its tokens is computed as:
\begin{equation}
\begin{aligned}
    A^{\text{matched}}_{\text{index}(j-1)<k\le \text{index}(j)} &= r_{\text{index}(j)} + r_{\text{index}(j+1)} + \cdots + r_{\text{index}(n)} \\ 
    &= \underbrace{\hat{r}^o + \cdots + \hat{r}^o}_{(n-j)/2 \text{ terms}} + \underbrace{(-\hat{r}^o) + \cdots + (-\hat{r}^o)}_{(n-j)/2 \text{ terms}} + \hat{r}^o \\
    &= \hat{r}^o.
\end{aligned}
\end{equation}

% \begin{equation}
% \begin{aligned}
% A^{\text{matched}}_{\text{index}(j-1)<k\le \text{index}(j)} &= r_{\text{index}(j)} + r_{\text{index}(j+1)} + \cdots + r_{\text{index}(n)} \\
% &= \underbrace{\hat{r}^o + \cdots + \hat{r}^o}_{(n-j)/2 \text{ terms}} \\
% & \qquad + \underbrace{(-\hat{r}^o) + \cdots + (-\hat{r}^o)}_{(n-j)/2 \text{ terms}} + \hat{r}^o \\
% &= \hat{r}^o.
% \end{aligned}
% \end{equation}

Similarly, for an unmatched thought $t_j^{\text{unmatched}}$, the advantage is:
\begin{equation}
    \begin{aligned}
        A^{\text{unmatched}}_{\text{index}(j-1)<k\le \text{index}(j)} &= r_{\text{index}(j)} + r_{\text{index}(j+1)} + \cdots + r_{\text{index}(n)} \\
        &= \underbrace{\hat{r}^o + \cdots + \hat{r}^o}_{(n-j+1)/2 - 1 \text{ terms}} + \underbrace{(-\hat{r}^o) + \cdots + (-\hat{r}^o)}_{(n-j+1)/2 \text{ terms}} + \hat{r}^o \\
        &= 0.
    \end{aligned}
\end{equation}

% \begin{equation}
% \begin{aligned}
% A^{\text{unmatched}}_{\text{index}(j-1)<k\le \text{index}(j)} &= r_{\text{index}(j)} + r_{\text{index}(j+1)} + \cdots + r_{\text{index}(n)} \\
% &= \underbrace{\hat{r}^o + \cdots + \hat{r}^o}_{(n-j+1)/2 - 1 \text{ terms}} \\
% & \qquad + \underbrace{(-\hat{r}^o) + \cdots + (-\hat{r}^o)}_{(n-j+1)/2 \text{ terms}} + \hat{r}^o \\
% &= 0.
% \end{aligned}
% \end{equation}

\end{proof}

\section{More Experiment Settings}\label{app:experiment_settings}
\subsection{Training Settings}
During process evaluation, we employ Qwen3-32B as the default evaluator. Due to the high computational cost of process evaluation, we implement some heuristic rules to skip some solutions that are unlikely to obtain process rewards: 
\begin{itemize}[leftmargin=0.5cm]
\item For solutions with correct answers, if the \textit{think} is short (fewer than 4,096 characters), we skip directly, as such \textit{think} typically contains only correct steps.
\item For solutions with incorrect answers, if the \textit{answer} is very short (fewer than 256 characters), we skip directly, as these \textit{answer} rarely contains obvious reasoning steps.
\item Solutions that are truncated due to length constraints are also skipped.
\end{itemize}

In the implementation of TP-GRPO and all replicated baseline experiments, we select GRPO as the base algorithm, and incorporate two modifications proposed in DAPO: $\epsilon_{high}$ and token-level loss. The outcome reward function combines accuracy and format rewards. 
% The accuracy reward is set as a binary accuracy reward (1 for correct answers, 0 otherwise). 
We extract the answer from `$\backslash\text{boxed}\{\}$' and compare it against the golden answer: a perfect match yields accuracy reward=1, otherwise 0. 
Additionally, we observed collapse when reproducing the on-policy GRPO baseline with $\beta = 0$. Therefore, following the open-r1 implementation~\citep{openr1}, we set $\beta = 0.04$ for the GRPO baseline. 
Additional learning-related hyperparameters and configurations are summarized in Table~\ref{tab:training_settings}.

\subsection{Evaluation Settings}
All evaluation experiments—including those for TP-GRPO and all baselines—are conducted using the evaluation codes from the \textbf{rllm} project~\footnote{ rllm: \url{https://github.com/agentica-project/rllm}}.

\begin{table}[!t]
    \centering
    % \resizebox{\linewidth}{!}{
    \begin{tabular}{ll}
    \toprule
    Hyperparameter & Value \\
    \midrule
    \textit{Sampling Configuration} & \\
    system prompt & \{problem\} Let's think step by step and \\
    & output the final answer within $\backslash\text{boxed}\{\}$. \\
    temperature & 1.0 \\ 
    max prompt length & 1024 \\
    max response length & 8192 \\
    rollout size & 8 \\
    \midrule
    \multicolumn{2}{l}{\textit{Process Evaluation Configuration}} \\
    default model & Qwen3-32B \\
    backend & VLLM \\
    temperature & 0.7 \\
    max tokens & 8192 \\
    % top P & 0.8 \\
    % top K & 20 \\
    % presence penalty & 1.5 \\
    tensor-parallel-size(VLLM) & 2 \\
    max\_num\_seqs(VLLM) & 512 \\
    \midrule
    \textit{Training Configuration} & \\
    gradient accumulation steps & 1 \\
    learning rate & 1.0e-06 \\
    lr scheduler type & constant \\
    max steps & 50 \\
    max iteration num & 20 \\
    per device train batch size & 5 \\
    process reward coefficient $\alpha$ & 1 \\
    $\epsilon_{low}$ & 0.2 \\
    $\epsilon_{high}$ & 0.28 \\
    \midrule
    \textit{Evaluation Configuration} & \\
    max tokens & 32768 \\
    temperature & 0.6 \\
    top P & 0.95 \\
    top K & -1 \\
    gpu\_memory\_utilization & 1.0 \\
    \bottomrule
    \end{tabular}
    % }
    \vspace{8pt}
    \caption{All key configuration hyperparameters used in this paper.}
    \label{tab:training_settings}
\end{table}
% \input{Tables/training_time}
% \section{More Experiments Results}

\section{More Experiments Results}

% \subsection{Training Time Analysis}
% Due to the high computational cost of GenPRM-based process evaluation, we adopt an off-policy training pipeline to improve efficiency. To validate its effectiveness, we compare not only training steps but also training time. As shown in Table~\ref{tab:training_time}, TP-GRPO achieves significantly shorter training time than the on-policy GRPO implementation based on the TRL framework.

% This improvement stems from TP-GRPO’s decoupling of sampling and optimization. During sampling, all GPU resources are dedicated to generating solutions; once completed, the GPUs are released and subsequently fully utilized for LRM training. In contrast, the TRL framework statically partitions GPUs resources: \textit{P} GPUs for sampling and \textit{Q} GPUs for training. During sampling, the \textit{Q} training GPUs remain idle; during optimization, the \textit{P} sampling GPUs sit unused. This underutilization of hardware resources leads to lower throughput and longer training time.

% \subsection{Hyperparameter Analysis}

\subsection{Case Studies}
To intuitively compare the improvements of our proposed TP-GRPO algorithm over the original GRPO, we selected one problem each from the AIME24 and AIME25. These problems were then presented to the DeepSeek-R1-Distill-Qwen-1.5B and DeepSeek-R1-Distill-Qwen-7B models, which were trained using both algorithms respectively, to observe their corresponding reasoning processes. The detailed results for the DeepSeek-R1-Distill-Qwen-1.5B model are shown in Table ~\ref{tab:1.5B_GRPO} and Table ~\ref{tab:1.5B_TP_GRPO}.
% , while the results for the DeepSeek-R1-Distill-Qwen-7B model are presented in Table XX and Table XX. 
As observed from the model's thinking content, the model post-trained with the TP-GRPO algorithm can correct the specific reasoning errors made by the model trained with the original GRPO. This provides an intuitive demonstration of the effectiveness of our proposed algorithm.
% \begin{tcolorbox}[colback=blue!10!white, colframe=blue!80!black, 
%   title=1.5B Model with GRPO Case Study, fonttitle=\bfseries]

\begin{tcolorbox}[colback=white, colframe=black, 
  title=1.5B Model with GRPO Case Study, fonttitle=\bfseries, breakable]

\textbf{[Question]}\\
Alice chooses a set $A$ of positive integers. Then Bob lists all finite nonempty sets $B$ of positive integers with the property that the maximum element of $B$ belongs to $A$. Bob's list has 2024 sets. Find the sum of the elements of A. Let's think step by step and output the final answer within \verb|\boxed{}|. \\

\textbf{[Model Thinking Content]}

Okay, so Alice picks a set A of positive integers. Then Bob lists all finite nonempty sets B of positive integers where the maximum element of B is in A. The total number of such sets B is 2024. We need to find the sum of the elements of A.

...\\

So, if A has n elements, then the total number of such sets B is the sum from k=1 to n of $2^{k-1}$. Because for each element $a_k$ in A, the number of subsets where $a_k$ is the maximum is $2^{k-1}$.\\
Therefore, the total number of sets B is $2^{n} - 1$, since the sum from k=0 to n-1 of $2^k$ is $2^n - 1$
So, if A has elements $a_1$, $a_2$, ..., 
$a_n$, then the total number is 
$2^{a_1 -1} + 2^{a_2 -1} + ... + 
2^{a_n -1} = 2024$

...\\
\mrc{Now, 2024 can be expressed as a sum of powers of 2. Let's see, 2024 in binary is 11111101100, which is 1024 + 512 + 256 + 128 + 64 + 32 + 4. Wait, but let me compute that.}\\
\mrc{Wait, 1024 is $2^10$, 512 is $2^9$, 256 is $2^8$, 128 is $2^7$, 64 is $2^6$, 32 is $2^5$, 16 is $2^4$, 8 is $2^3$, 4 is $2^2$, 2 is $2^1$, 1 is $2^0$. So, 1024 + 512 is 1536, plus 256 is 1792, plus 128 is 1920, plus 64 is 1984, plus 32 is 2016, plus 4 is 2020.
Wait, that's $1024+512+256+128+64+32+4=2020$. Hmm,
but 2024 is 4 more, so 2020 + 4 is 
2024. So, in terms of powers of 2, 
$2024 = 2^{10} + 2^9 + 2^8 + 2^7 + 2^6
+ 2^5 + 2^2$.}

...\\
\mrc{So, the exponents $a_i -1$ are 2, 5, 6, 7,
8, 9, 10.
Therefore, $a_i$ are 3,6,7,8,9,10,11.
Therefore, the set A is 3,6,7,8,9,10,11.
Therefore, the sum of the elements 
of A is 3 + 6 + 7 + 8 + 9 + 10 + 11.}

...\\
\mrc{Therefore, the sum of the elements of A is 3 + 6 + 7 + 8 + 9 + 10 + 11.\\
Let me compute that:\\
3 + 6 is 9.\\
9 +7 is 16.\\
16 +8 is 24.\\
24 +9 is 33.\\
33 +10 is 43.\\
43 +11 is 54.\\
So, the sum is 54.\\
Therefore, the answer is 54. \\
\textbf{Final Answer}: The sum of the 
elements of A is $\boxed{54}$.}

\end{tcolorbox}

\captionof{table}{Case study of DeepSeek-R1-Distill-Qwen-1.5B with GRPO. For the given problem, the model incorrectly answered 54, while the correct answer is 55. Due to the model's extensive thinking process, we have excerpted the key steps for clarity. As shown, the error (highlighted by red) stems from a miscalculation in the binary representation of 2024.}
\label{tab:1.5B_GRPO}
\begin{tcolorbox}[colback=white, colframe=black, 
  title=1.5B Model with TP-GRPO Case Study, fonttitle=\bfseries, breakable]

\textbf{[Question]}\\
Alice chooses a set $A$ of positive integers. Then Bob lists all finite nonempty sets $B$ of positive integers with the property that the maximum element of $B$ belongs to $A$. Bob's list has 2024 sets. Find the sum of the elements of A. Let's think step by step and output the final answer within \verb|\boxed{}|. \\

\textbf{[Model Thinking Content]}

Okay, so I have this problem here where Alice chooses a set A of positive integers. Then Bob lists all finite nonempty sets B of positive integers with the property that the maximum element of B belongs to A. Bob's list has 2024 sets. I need to find the sum of the elements of A. Hmm, let me try to break this down step by step.

...\\
So for each a in A, the number of such sets is equal to the number of subsets of {1, 2, ..., a-1}. Because each subset can be combined with a to form a set B with maximum a.\\
Therefore, for each a in A, the number of sets B with maximum a is $2^{a-1}$. Because the number of subsets of a set with n elements is $2^n$. So in this case, n is a-1, so it's $2^{a-1}$.\\
So, if A is a set of numbers, say $A = {a_1, a_2, ..., a_k}$, then the total number of sets B that Bob can list is the sum over each $a_i$ in A of $2^{a_i - 1}$. Because each $a_i$ contributes $2^{a_i - 1}$ sets. Therefore, the total number of sets B is the sum of $2^{a_i - 1}$ for all $a_i$ in A. So, let me denote that $S = \sum_a^A 2^{a - 1} = 2024$. 

...\\

So, let's factor 2024 into its prime factors to see if we can represent it as a sum of powers of 2.\\
First, 2024 divided by 2 is 1012, divided by 2 again is 506, divided by 2 again is 253. 253 is not divisible by 2, so the prime factors are $2^3 \times 253$.\\
Now, 253 divided by 11 is 23, because 11*23 is 253. So, $2024 = 2^3 \times 11 \times 23$. So, $2024 = 8 \times 11 \times 23$. Hmm, interesting.\\
But how does this help me? I need to express 2024 as the sum of $2^{a - 1}$ terms, each of which is a power of 2, and each a is a positive integer. So, each term is a power of 2, and the sum is 2024. So, to find A, I need to express 2024 as the sum of distinct powers of 2

...\\

\rit{So, let me compute the binary 
representation of 2024. Calculating:\\
$2^{10} = 1024$\\
$2^9 = 512$\\
$2^8 = 256$\\
$2^7 = 128$\\
$2^6 = 64$\\
$2^5 = 32$\\
$2^3 = 8$\\
So, summing these up:\\
1024 + 512 = 1536\\
1536 + 256 = 1792\\}
\rit{1792 + 128 = 1920\\
1920 + 64 = 1984\\
1984 + 32 = 2016\\
2016 + 8 = 2024\\
Yes, that adds up correctly. So, 2024 in
binary is 11111101000.
So, the binary representation has 1s at 
positions 10,9,8,7,6,5,3. So, 
that would mean that the exponents 
in the sum are 10,9,8,7,6,5,3.}

...\\

\rit{So, if the exponents are 10,9,8,7,6,5,3, then the corresponding a's are 11,10,9,8,7,6,4.\\
Therefore, set A is {4,6,7,8,9,10,11}.\\
So, the sum of elements in A is 4 + 6 + 7 + 8 + 9 + 10 + 11.\\
Let me compute that:\\
4 + 6 = 10\\
10 + 7 = 17\\
17 + 8 = 25\\
25 + 9 = 34\\
34 + 10 = 44\\
44 + 11 = 55\\
So, the sum is 55.}

...\\

\rit{\textbf{Final Answer}: The sum of the 
elements of A is $\boxed{55}$.}

\end{tcolorbox}

\captionof{table}{Case study of DeepSeek-R1-Distill-Qwen-1.5B with TP-GRPO. Using the same problem from Table ~\ref{tab:1.5B_GRPO}, we also present only the key steps. Post-training with TP-GRPO, the model successfully computes the binary representation of 2024 and yields the correct final answer. We use blue to highlight these steps.}
\label{tab:1.5B_TP_GRPO}
\begin{tcolorbox}[colback=white, colframe=black, 
  title=7B Model with GRPO Case Study, fonttitle=\bfseries, breakable]

\textbf{[Question]}\\
Let $ A $ be the set of positive integer divisors of 2025. Let $ B $ be a randomly selected subset of $ A $. The probability that $ B $ is a nonempty set with the property that the least common multiple of its elements is 2025 is $\frac{m}{n}$, where $ m $ and $ n $ are relatively prime positive integers. Find $ m + n $. Let's think step by step and output the final answer within \verb|\boxed{}|.\\

\textbf{[Model Thinking Content]}

Okay, so I have this problem here where I need to find the probability that a randomly selected subset B of the set A (which consists of all positive integer divisors of 2025) has the property that the least common multiple (LCM) of its elements is 2025. 

...\\
I know that 2025 divided by 25 is 81, because 25 times 81 is 2025. And 81 is 9 squared, so 2025 is 25 times 81, which is 5 squared times 3 to the fourth power. So, $2025 = 5^2 × 3^4$. Therefore, the prime factorization of 2025 is $5^2 × 3^4$.\\
Now, to find all the positive divisors of 2025, I can use the exponents of its prime factors. The number of divisors is given by multiplying one more than each of the exponents in the prime factorization. So here, for $5^2$, the exponent is 2, so we have 2 + 1 = 3. For $3^4$, the exponent is 4, so 4 + 1 = 5. Therefore, the total number of divisors is 3 × 5 = 15. So, set A has 15 elements.

...\\
Next, we need to consider all possible subsets B of A, and find the probability that the LCM of the elements in B is 2025. Since we'{}re dealing with subsets, the total number of subsets is $2^{15}$, because each element can either be included or excluded. However, the problem specifies that B is a nonempty set, so we have to subtract 1 for the empty set. \mrc{Therefore, the total number of possible nonempty subsets is $2^{15} - 1$.}\\
\mrc{Wait, actually, let me think. The problem says ``a randomly selected subset of A." It doesn'{}t specify whether it\'s nonempty or not. But then, it goes on to specify that B is a nonempty set with the property that LCM of its elements is 2025. So, perhaps the probability is considering all subsets, including the empty set, but the event we'{}re interested in is only when B is nonempty. Hmm, the wording is a bit ambiguous.\\
Looking back at the problem: ``The probability that B is a nonempty set with the property that the least common multiple of its elements is 2025." So, it'{}s given that B is a nonempty set, so we can consider the total number of subsets as $2^{15} - 1$, since we exclude the empty set. Therefore, the probability is $\frac{(number\ of\ desired \ subsets)}{(2^{15} - 1)}$.}

...\\
Given that $2025 = 3^4 × 5^2$, the LCM 
of a subset B is 2025 if and only if
the subset contains elements that 
cover all the prime factors with 
their maximum exponents.

...\\
By inclusion-exclusion, this number is $U - |A| - |B| + |A \cap B|$.\\
\mrc{So, $|A|$ is the number of subsets without 81. Since 81 is one specific element, the number of subsets without 81 is $2^{14}$.\\
Similarly, $|B|$ is the number of subsets without 25, which is also $2^{14}$.\\
$|A \cap B|$ is the number of subsets without both 81 and 25, which is $2^{13}$.\\
Therefore, the number of subsets containing at least one 81 and at least one 25 is:\\
$2^{15} - 2^{14} - 2^{14} + 2^{13}$.\\
Let me compute this:\\
$2^{15} = 32768$\\
$2^{14} = 16384$\\
$2^{13} = 8192$\\}
\mrc{So, 32768 - 16384 - 16384 + 8192.\\
Compute step by step:\\
32768 - 16384 = 16384\\
16384 - 16384 = 0\\
0 + 8192 = 8192\\
So, the number of subsets containing at least one 81 and at least one 25 is 8192.}

...\\
\mrc{Therefore, the probability is $\frac{8191}{(2^{15} - 1)} = \frac{8191}{32767}$}.\\
Now, we need to simplify this fraction. Let'{}s see if 8191 and 32767 have any common factors.\\
First, let me factorize 32767. I know that 32767 is $2^{15} - 1$, which is 32768 - 1 = 32767.\\
32767 is a Mersenne number, $2^{15} - 1$. Let'{}s see if it'{}s prime.\\
Well, $2^{15}$ is 32768, so 32767 is 32768 - 1. Let me check its divisibility.\\
$32767 ÷ 7: 7 × 4681 = 32767$? Let'{}s check: 7 × 4000 = 28000, 7 × 681 = 4767, so 28000 + 4767 = 32767. Yes, so 32767 = 7 × 4681.\\
Now, let'{}s factor 4681. Let me check if 4681 is divisible by 7: 7 × 668 = 4676, 4681 - 4676 = 5, so no. Next, 11: 4 - 6 + 8 - 1 = 5, which is not divisible by 11. 13: 4681 ÷ 13. 13 × 360 = 4680, so 4681 - 4680 = 1, so no. 17: 17 × 275 = 4675, 4681 - 4675 = 6, not divisible. 19: 19 × 246 = 4674, 4681 - 4674 = 7, no. 23: 23 × 203 = 4669, 4681 - 4669 = 12, no. 29: 29 × 161 = 4669, same as above. 31: 31 × 151 = 4681? Let'{}s compute 31 × 150 = 4650, plus 31 = 4681. Yes! So, 4681 = 31 × 151.\\
Therefore, 32767 = 7 × 31 × 151.\\
Now, let'{}s check if 8191 is divisible by any of these primes: 7, 31, 151.\\
8191 ÷ 7: 7 × 1170 = 8190, so 8191 - 8190 = 1, so remainder 1. Not divisible by 7.\\
8191 ÷ 31: Let'{}s see, 31 × 264 = 8184, 8191 - 8184 = 7, so remainder 7. Not divisible by 31.\\
8191 ÷ 151: 151 × 54 = 8154, 8191 - 8154 = 37, which is less than 151, so not divisible.\\
Therefore, 8191 and 32767 share no common factors besides 1, so the fraction $\frac{8191}{32767}$ is already in its simplest form.

...\\
\mrc{Therefore, m = 8191, n = 32767, and 
m + n = 40958.\\
I think that'{}s solid. I don'{}t see any mistakes in the reasoning.\\
\textbf{Final Answer}: $\boxed{40958}$}

\end{tcolorbox}

\captionof{table}{Case study of DeepSeek-R1-Distill-Qwen-7B with GRPO. For the given problem, the model incorrectly answered 40958, while the correct answer is 237. Due to the model's extensive thinking process, we have excerpted the key steps for clarity. As shown, the error (highlighted by red) mainly from a miscalculation in the number of corresponding subsets.}

\label{tab:7B_GRPO}
\begin{tcolorbox}[colback=white, colframe=black, 
  title=7B Model with TP-GRPO Case Study, fonttitle=\bfseries, breakable]

\textbf{[Question]}\\
Let $ A $ be the set of positive integer divisors of 2025. Let $ B $ be a randomly selected subset of $ A $. The probability that $ B $ is a nonempty set with the property that the least common multiple of its elements is 2025 is $\frac{m}{n}$, where $ m $ and $ n $ are relatively prime positive integers. Find $ m + n $. Let's think step by step and output the final answer within \verb|\boxed{}|. \\

\textbf{[Model Thinking Content]}

Okay, so I have this problem here where I need to find the probability that a randomly selected subset B of the set A (which consists of all positive integer divisors of 2025) has the property that the least common multiple (LCM) of its elements is 2025. \\
...\\
A is the set of all positive integer divisors of 2025. So, let me factorize 2025 to find out its divisors. 2025 is a number I might recognize as a square. Let me check: 45 squared is 2025 because $45 \times 45$ is 2025. So, 2025 = $45^2$. But 45 itself can be broken down further: $45 = 9\times5$, and 9 is $3^2$. So, 2025 is $3^4 \times 5^2$. Therefore, the prime factorization of 2025 is $3^4 \times 5^2$.\\
Now, to find the number of positive divisors, we can use the formula for the number of divisors based on prime factorization. If a number N has the prime factorization $N = p_1^{a_1} * p_2^{a_2} * ... * p_n^{a_n}$, then the number of positive divisors is $(a_1 + 1)(a_2 + 1)...(a_n + 1)$. So, for 2025, which is $3^4 \times 5^2$, the number of divisors is $(4 + 1)\times(2 + 1) = 5\times3 = 15$. So, set A has 15 elements.\\
...\\
Next, I need to consider the power set of A, which is all possible subsets of A, including the empty set and A itself. The number of subsets of a set with n elements is $2^n$. \rit{Since A has 15 elements, the power} \rit{set has $2^{15}$ subsets. Therefore, the total number of} \rit{possible subsets B is 32768.}\\
Now, the problem is asking for the probability that a randomly selected subset B has an LCM of 2025. So, I need to find how many such subsets exist where the LCM of their elements is 2025, and then divide that by the total number of subsets, which is 32768.\\
...\\
Given that 2025 is $3^4 \times 5^2$, the LCM of a subset will be 2025 if and only if in the subset, there is at least one number with $3^4$ and at least one number with $5^2$. Wait, is that correct? Or is it sufficient that the maximum exponent of 3 in the subset is 4 and the maximum exponent of 5 is 2?\\
Yes, exactly. So, for the LCM to be 2025, the subset must include numbers such that the highest power of 3 in the subset is $3^4$ and the highest power of 5 is $5^2$. So, in other words, the subset must contain at least one multiple of $3^4$ (which is 81) and at least one multiple of $5^2$ (which is 25). 

...\\

Therefore, the number of subsets with LCM 2025 is equal to the number of subsets that include at least one multiple of 81 and at least one multiple of 25. So, to compute this, we can use the principle of inclusion-exclusion.\\
First, let's denote:\\
- Let S be the set A, which has 15 elements.\\
- Let X be the set of subsets of S that include at least one multiple of 81.\\
- Let Y be the set of subsets of S that include at least one multiple of 25.\\
We need to find $|X \cap Y|$, which is the number of subsets that include at least one multiple of 81 and at least one multiple of 25.\\
By the principle of inclusion-exclusion:\\
$|X \cap Y| = |X| + |Y| - |X \cup Y|$
So, let's compute each term.\\
Total subsets: $2^{15}$.\\
\rit{Subsets without any multiple of 81: So, we need to count the number of subsets that do not include 81. 2025 is $3^4 \times 5^2$, so the divisors are of the form $3^a \times 5^b$ where a is 0,1,2,3,4 and b is 0,1,2.\\
The number of divisors is 15. The number of divisors that are multiples of 81: 81 is $3^4$, so any multiple of 81 must have a=4. So, the number of such divisors is equal to the number of possible b's, which are 0,1,2. So, there are 3 divisors that are multiples of 81: $81 \times 1 = 81, 81 \times 5 = 405, 81 \times 25 = 2025.$}\\
\rit{Therefore, the number of elements not divisible by 81 is 15 - 3 = 12. Therefore, the number of subsets without any multiple of 81 is $2^{12}$.\\
Similarly, subsets without any multiple of 25: 25 is $5^2$. So, the divisors that are multiples of 25 are those with b=2. The number of such divisors is equal to the number of a's, which are 0,1,2,3,4.} \rit{So, that's 5 divisors: 1, 5, 25, 75, 225, 2025? Wait, no: 1 is $3^0 \times 5^0$, 5 is $3^0 \times 5^1$, 25 is $3^0 \times 5^2$, 75 is $3^1 \times 5^2$, 225 is $3^2 \times 5^2$, 2025 is $3^4 \times 5^2$. Wait, that's 6 divisors. Wait, but 2025 is $3^4 \times 5^2$, so for b=2, a can be 0,1,2,3,4: that's 5 divisors. Wait, $3^0 \times 5^2, 3^1 \times 5^2, ..., 3^4 \times 5^2$: that's 5 divisors. Wait, why did I think 6? Hmm, maybe I confused with 2025's total divisors.\\
Wait, no, the number of divisors where b=2 is 5 because a can be 0,1,2,3,4. So, 5 divisors. Therefore, the number of divisors not divisible by 25 is 15 - 5 = 10. Therefore, the number of subsets without any multiple of 25 is $2^{10}$.\\
Now, the subsets that include neither any multiple of 81 nor any multiple of 25: So, these are subsets that do not contain any of the 3 multiples of 81 and do not contain any of the 5 multiples of 25. So, how many such elements are there?\\
Total elements:15.\\}
\rit{Number of elements that are multiples of 81 or 25: Let's compute this using inclusion-exclusion. Number of multiples of 81:3, multiples of 25:5, and multiples of both 81 and 25: Well, 81 and 25 are coprime, so their least common multiple is 81 x 25 = 2025. So, only 2025 is a multiple of both 81 and 25. Therefore, the number of elements that are multiples of 81 or 25 is 3 + 5 -1 =7.\\
Therefore, the number of elements that are neither multiples of 81 nor multiples of 25 is 15  - 7 = 8.\\
Therefore, the number of subsets that include neither any multiple of 81 nor any multiple of 25 is $2^8$.\\
Therefore, putting it all together:\\
$|X \cap Y|$ = Total subsets - subsets without 81 - subsets without 25 + subsets without both 81 and 25\\
Which is:\\
$|X \cap Y| = 2^{15} - 2^{12} - 2^{10} + 2^8$\\
Let me compute these values step by step.\\
First, $2^{15} = 32768$\\
$2^{12} = 4096$\\
$2^{10} = 1024$\\
$2^8 = 256$\\
So, substituting:\\
$|X \cap Y| = 32768 - 4096 - 1024 + 256$\\
Let me compute this step by step:\\
First, 32768 - 4096 = 28672\\
Then, 28672 - 1024 = 27648\\
Then, 27648 + 256 = 27904\\
So, $|X \cap Y| = 27904$ \\
Therefore, the number of subsets B where LCM(B) = 2025 is 27904.}\\
...\\

\rit{Therefore, the probability is $\frac{27904}{32768}$}.\\
Now, we need to reduce this fraction to its lowest terms. Let's find the greatest common divisor (GCD) of 27904 and 32768.\\
First, let's note that both numbers are divisible by 16: 27904 ÷ 16 = 1744, 32768 ÷ 16 = 2048.\\
So, 1744 and 2048.\\
Again, both are divisible by 16: 1744 ÷ 16 = 109, 2048  ÷ 16 = 128.\\
So, 109 and 128. Now, 109 is a prime number (since it's not divisible by 2,3,5,7,11; $\sqrt{109} \approx 10.4$, so primes up to 11 are tested). 128 is $2^7$. Since 109 is prime and doesn't divide 128, the GCD is 1.\\
Therefore, the reduced fraction is $\frac{1744}{2048}$. Wait, no: wait, 27904 ÷ 16 = 1744, 32768 ÷ 16 = 2048. Then, 1744 ÷ 16 = 109, 2048 ÷ 16 = 128. So, the reduced fraction is $\frac{109}{128}$.\\
Wait, but 1744 ÷16=109? Wait, 16*109=1744? Let me compute 16*100=1600, 16*9=144, so 1600+144=1744. Yes, correct. Similarly, 16*128=2048, which is correct.\\
So, 109 and 128 are coprime, so the reduced fraction is $\frac{109}{128}$.\\
Therefore, the probability is $\frac{109}{128}$, so m=109, n=128.\\
Thus, m + n=109 + 128 = 237.\\
Wait, hold on, but let me confirm the calculations once more because 27904 divided by 32768 reduces by 16 twice, leading to $\frac{109}{128}$, which is correct.\\
...\\

\rit{Therefore, the probability is $\frac{27904}{32768} = \frac{109}{128}$, so m + n = 237.\\
Thus, I think that is the correct answer.\\
\textbf{Final Answer}: \ $\boxed{237}$}

\end{tcolorbox}

\captionof{table}{Case study of DeepSeek-R1-Distill-Qwen-7B with TP-GRPO. Using the same problem from Table ~\ref{tab:7B_GRPO}, we also present only the key steps. Post-training with TP-GRPO, the model successfully computes the number of corresponding subsets and yields the correct final answer. We use blue to highlight these steps.}
\label{tab:7B_TP_GRPO}

\section{Process Evaluation Details}\label{app:prompt_details}
In this part, we will provide the implementation details of the process evaluation with GenPRM.
\paragraph{Atomic Logical Actions-Based Decompose Details}
In our process, a solution must first be decomposed into a series of steps with appropriate granularity for subsequent processing. However, since solutions generated by the model are often lengthy, it is difficult to directly decompose them into steps that are both content-complete and have clear boundaries. Here, we first split each solution into shorter sentences based on the "$\backslash$n$\backslash$n" delimiter. However, we observed in practice that the model often generates "$\backslash$n$\backslash$n" in the middle of a process, leading to severe fragmentation of the resulting sentences and a vast number of segments. Therefore, through observation we defined a list of punctuation marks that signify the end of the finest-grained meaningful sentences. Based on this list, we sequencialy merge consecutive short sentences until the current sentence terminates with one of these specified punctuation marks. The punctuation marks list is shown in Table ~\ref{tab:punctuation_marks_list}. 
% as follows: $[.,..., ...., ....., ......, !, ?]$.
To facilitate subsequent processing, we then insert a sequentially increasing label (e.g., \textless\textless0\textgreater\textgreater, \textless\textless1\textgreater\textgreater, ...) at the beginning of each merged sentence. To ensure that each sentence has an appropriate semantic granularity, we define eight atomic logical actions and prompt GenPRM to further merge consecutive sentences based on these actions. Ultimately, for each solution we obtain a series of steps with appropriate quantity and granularity, as well as clear semantic boundaries. The prompt used is shown in Table ~\ref{tab:solution_decompose_prompt}. For incorrect solutions, we also need to decompose their corresponding answers. Since answers are typically much shorter than solutions, we directly decompose it using the atomic logical actions mentioned above, bypassing the preprocessing stage. Specifically, we instruct GenPRM to enclose each delineated step within "\textless step\textgreater" and "\textless/step\textgreater" tags. It facilitates the subsequent parsing of steps using regular expressions, and mitigates the risk of GenPRM corrupting the original answer (e.g., by rewriting or omitting parts). The prompt used is shown in Tabel ~\ref{tab:answer_decompose_prompt}.

\paragraph{Subsequent Match-Based Step Type Check for Incorrect Solutions Details}
For solutions that remain excessively long (i.e., exceeding 8192 characters), after the initial decomposition we apply an additional step-level summarization using GenPRM. This makes them more concise for subsequent processing. The prompt used is shown in Table ~\ref{tab:summarize_solution_prompt}. Subsequently, to match each step of the answer to the corresponding steps in the solution, we focus solely on semantic and content similarity, disregarding whether the solution steps themselves are correct or not. The prompt used can be found in Table ~\ref{tab:match_prompt}.

\paragraph{Subsequent Reflection-Based Step Type Check for Correct Solutions Details}
For correct solutions, we first provide the step-wise decomposed solution to GenPRM, tasking it with identifying all reflection segments. However, solution may contains validation-style reflections (e.g., ``double-checking") that do not identify specific errors. We constrain the model to only find reflection segments that point out concrete errors. The prompt used is shown in Table ~\ref{tab:aware_reflection_prompt}. Next, we iterate through each of the identified reflection segments to trace its error source, which we define as the earliest step where the error mentioned in the reflection first appeared. The prompt used in shown in Table ~\ref{tab:aware_mistake_source_prompt}. After identifying the error source for each reflection segment, we retain only the source with the minimum index, as this is crucial for meaningfully identifying the ultimate origin of the error. Specifically, if multiple reflection segments map to the earliest source, we keep the reflection segment with the maximum index. Based on the principle of semantic consistency, all reflections within this span are considered semantically related to the source error, either directly or as an indirect consequence. Finally, we task GenPRM to determine the type of each step within it. Specifically, we define three types: \textbf{Incorrect}, \textbf{Reflection}, and \textbf{Correct}.
\begin{itemize}[leftmargin=1cm]
\item \textbf{Incorrect:} An Incorrect step is one that contains a logical or computational error.
\item \textbf{Reflection:} A Reflection step is one that contains a reflective action.
\item \textbf{Correct:} Steps that fall into neither of the two aforementioned types.
\end{itemize}
The prompt used for this classification task is shown in Table ~\ref{tab:reason_all_mistake_prompt}. As our overall workflow is divided into three stages, the actual implementation involves parsing the JSON data returned from the prompt of the preceding stage, and then passing to the prompt for the subsequent stage.

\begin{table}[htbp] % htbp 是建议的浮动位置选项
    \centering % 让盒子在页面上居中显示

    % --------------------------
    \begin{tcolorbox}[colback=orange!10!white, colframe=orange!80!black,  title=Punctuation Marks List, fonttitle=\bfseries]

    \begin{verbatim}
[., ..., ...., ....., ......, !, ?]
    \end{verbatim}

    \end{tcolorbox}
    % -------------------------

    % --- 在这里添加标题和标签 ---
    \caption{The punctuation marks list we use to merge consecutive pre-split sentences.}
    \label{tab:punctuation_marks_list} % 用于交叉引用的标签
\end{table}

\section{More details on Evaluation Prompts}
\begin{table}[htbp] % htbp 是建议的浮动位置选项
    \centering % 让盒子在页面上居中显示
    
    % 使用 \input 导入您的 tcolorbox 文件
    \begin{tcolorbox}[colback=orange!10!white, colframe=orange!80!black, 
  title=Summarize Solution Prompt, fonttitle=\bfseries]

You will be presented with the following two contents:\\

\textbf{[Math Problem]}\\
\{problem\}\\

\textbf{[Solution]}\\
\{solution in a list format\}\\

This thought process includes \textbf{$len(solution)$} Steps in total.\\
Your task is to summarize the main content of these all Steps in the solution. \\
Make sure each summary clearly describes the key points and the intermediate conclusion of the respective step in the thought process.\\

Finally, your output should be in standard \textbf{JSON} format:\\
\textasciigrave \textasciigrave \textasciigrave json
\begin{verbatim}
{'Thought Step 0': 'Summary of Step 
0', 'Thought Step 1': 'Summary of 
Step 1', ...}
\end{verbatim}
\textasciigrave \textasciigrave \textasciigrave

\end{tcolorbox}
    
    % --- 在这里添加标题和标签 ---
    \caption{The prompt used to summarize each step of a solution. Note that the input solution has already been decomposed into steps in a list format.}
    \label{tab:summarize_solution_prompt} % 用于交叉引用的标签
\end{table}
\begin{table}[htbp] % htbp 是建议的浮动位置选项
    \centering % 让盒子在页面上居中显示
    
    % 使用 \input 导入您的 tcolorbox 文件
    \begin{tcolorbox}[colback=orange!10!white, colframe=orange!80!black, 
  title=Solution Decompose Prompt, fonttitle=\bfseries]
You need to read the problem-solving process below, which is divided into some steps using line breaks, with each step labeled by a tag “\textless\textless\textgreater\textgreater” .\\

\textbf{[Pre Split Format Thought]}\\
\{format solution\}\\

Then, you are required to combine consecutive steps into several logically independent sections. Note that each independent section must belong to one of the eight action categories listed below:\\

(1) Problem Analysis: Decompose the original problem, clarify known conditions and objectives\\
(2) Subtask Definition: Establish phased goals (e.g., equation construction, variable definition)\\
(3) Calculation Solving: Perform numerical operations or symbolic derivations\\
(4) Reflection Check: Verify logical/computational errors\\
(5) Answer Validation: Confirm correctness of final/intermediate conclusions\\
(6) Error Correction: Rectify identified issues\\
(7) Strategy Shift: Switch to alternative problem-solving approach\\
(8) Final Conclusion: Summarize and present ultimate answer\\

\textbf{**Output Specifications**}\\
- Maintain original thought chain order when outputting sections\\
- Section format: \textless\textless start \textgreater\textgreater - \textless\textless end \textgreater\textgreater \ Concise explanation the reasons for combining these steps together(max 15 words)\\
- Must satisfy: $start \le end < \text{next\_start}$\\
   
\textbf{**Special Constraints**}\\
- If several consecutive steps involving calculations, enumerations, proofs, or similar processes are working toward the same objective—such as solving the same equation, proposition, or problem—please group them together into a single section without splitting them apart!\\
- If several consecutive steps are reflections on or verifications for the same issue, combine them into a single section and do not separate them!\\

\textbf{**Example output**}\\
\textless\textless 0 \textgreater\textgreater \ - \textless\textless 2 \textgreater\textgreater  These three steps aim to analyze the question\\
\textless\textless 3 \textgreater\textgreater \ - \textless\textless 5 \textgreater\textgreater  These three steps is computing the same equation\\
\textless\textless 6 \textgreater\textgreater \ - \textless\textless 6 \textgreater\textgreater This step aims to draw the final answer

\end{tcolorbox}
    
    % --- 在这里添加标题和标签 ---
    \caption{The prompt used to decompose preprocessed solutions. We define eight atomic logical actions and use them to guide the step-wise decomposition.}
    \label{tab:solution_decompose_prompt} % 用于交叉引用的标签
\end{table}
\begin{table}[htbp] % htbp 是建议的浮动位置选项
    \centering % 让盒子在页面上居中显示
    
    % 使用 \input 导入您的 tcolorbox 文件
    \begin{tcolorbox}[colback=orange!10!white, colframe=orange!80!black, 
  title=Match Solution and Answer Prompt, fonttitle=\bfseries]

You will be given the following three sections:\\

\textbf{[Math Problem]}\\
\{problem\}\\
\textbf{[Summarized Solution]}\\
\{summarized solution in a list format\}\\
\textbf{[Answer]}\\
\{answer of the problem in a list format\}\\

\textbf{**Task Requirements**}\\
1. Establish a mapping relationship between all these solution steps and the answer steps.\\
2. For each answer step, find all solution steps whose summaries match the content of the current solution step summary.\\
3. Allow one-to-many or many-to-one correspondence.\\

\textbf{**Matching Criteria**}\\
1. The Answer Step should summarize the corresponding Solution Step, or be a refined articulation of it.\\
2. If the LAST step of the Solution includes the "final answer" from the Answer, it should at least match the LAST step of the Answer.\\

\textbf{**Output Requirements**}\\
1. Analysis Report:\\
  - Explain the matching basis for each solution step in order.\\
  - Only record clear content correspondences.\\
2. Mapping Table (\textbf{JSON} format):
\begin{verbatim}
{"Solution_Step_Number": 
[Thought_Step_Numbers], ...}
\end{verbatim}
  - Unmatched thought process steps are not included.\\
3. Example:\\
\textasciigrave \textasciigrave \textasciigrave json
\begin{verbatim}
{"0": [1, 3],"1": [5]} 
\end{verbatim}
\textasciigrave \textasciigrave \textasciigrave

\end{tcolorbox}
    
    % --- 在这里添加标题和标签 ---
    \caption{The prompt used to match each answer step with its corresponding solution step(s). Note that both the solution and the answer have already been decomposed into steps in a list format.}
    \label{tab:match_prompt} % 用于交叉引用的标签
\end{table}
\begin{table}[htbp] % htbp 是建议的浮动位置选项
    \centering % 让盒子在页面上居中显示
    
    % 使用 \input 导入您的 tcolorbox 文件
    \begin{tcolorbox}[colback=orange!10!white, colframe=orange!80!black, 
  title=Aware Reflection Prompt, fonttitle=\bfseries]

\textbf{**Instruction:**}\\
You will receive the following two pieces of content:\\
\textbf{[Math Problem]}\\
\{problem\}\\

\textbf{[Decomposed Solution]}\\
\{decomposed solution in a list format\}\\

Note that this thought process has been broken down into multiple steps, documenting the complete path from the initial attempt to the final correct solution. It may include attempts with incorrect reasoning, reflections on those attempts, and subsequent corrections.\\

\textbf{**Your Task:**}\\
Carefully read and analyze the provided thought process to identify all segments where 'an error in previous reasoning steps was discovered through reflection or verification.' For each identified segment, complete the following tasks:\\
1. Record the Step Index: Indicate the step number where this segment occurs.\\
2. Identify the Fragment: Clearly specify the exact content of these segments.\\
3. Summarize the Cause: Analyze and summarize the reason why the error occurred.\\

\textbf{**Output Format Requirements:**}\\
Return your results in the following \textbf{JSON} format:\\
\textasciigrave \textasciigrave \textasciigrave json
\begin{verbatim}
{"Reflection 1": {"Step Index": 
"The step number where this segment 
occurs", "Identified Fragment": 
"The exact content of the segment 
where an error in previous reasoning 
steps was discovered through 
reflection or verification", 
"Reason": "A summary of the reason 
why the error occurred"}, ...}
\end{verbatim}
\textasciigrave \textasciigrave \textasciigrave\\

\textbf{**Notes:**}\\
1. The steps you find MUST have identified a previous ERROR through reflection. If this step ONLY involves double-checking a previous calculation to verify its correctness, without finding a specific error, then ignore this step.\\
2. If there are no qualifying segments in the thought process, return an empty \textbf{JSON} object: `\{\}'.\\

\end{tcolorbox}
    
    % --- 在这里添加标题和标签 ---
    \caption{The prompt used to find reflection segments within a correct solution. Note that the input solution has already been decomposed into steps in a list format.}
    \label{tab:aware_reflection_prompt} % 用于交叉引用的标签
\end{table}
\begin{table}[htbp] % htbp 是建议的浮动位置选项
    \centering % 让盒子在页面上居中显示
    
    % 使用 \input 导入您的 tcolorbox 文件
    \begin{tcolorbox}[colback=orange!10!white, colframe=orange!80!black, 
  title=Aware Mistake Source Prompt, fonttitle=\bfseries]

 You will see a math problem and a pre-decomposed thought process for solving this math problem:\\
 \textbf{[Math Problem]}\\
 \{problem\}\\
 
\textbf{[Decomposed Solution]}\\
\{decomposed solution in a list format\}\\

This solution records the complete mental journey from initial attempts to the final correct answer, including trials of incorrect approaches, reflections, and corrections.\\

The following paragraph explicitly highlights the discovery of a mistake through reflection:\\
\textbf{[Identified Fragment]}\\
\{reflection identified fragment\}\\

And the reason of the mistake is summarized as following:\\
\textbf{[Reason]}\\
\{reflection reason\}\\

Your task is to identify the earliest step in the thought process that is responsible for this mistake.\\

\textbf{**Task Requirements**}\\
-The step you find MUST itself actually contain logical or calculation errors. You need to verify the step you have identified CAREFULLY.\\
- Before stating the earliest step you believe is responsible for this mistake, you need to carefully think and provide your detailed analysis.\\
- Finally, based on your analysis, output the earliest thought step and step number that you believe is responsible for the mistake, and provide your reason.\\

\textbf{**Output Format Requirements:**}\\
Return the final results in the following \textbf{JSON} format:\\
\textasciigrave \textasciigrave \textasciigrave json
\begin{verbatim}
{"Earliest Step": "The earliest step 
content that is responsible for this 
mistake", "Step Index": "The step 
number of this thought step", 
"Reason": "The reason you believe 
this step is responsible for this 
mistake"}
\end{verbatim}
\textasciigrave \textasciigrave \textasciigrave

\end{tcolorbox}
    
    % --- 在这里添加标题和标签 ---
    \caption{The prompt used to find the error source based on a reflection segment. This prompt utilizes information returned from the prompt in Table ~\ref{tab:aware_reflection_prompt}.}
    \label{tab:aware_mistake_source_prompt} % 用于交叉引用的标签
\end{table}
\begin{table}[htbp] % htbp 是建议的浮动位置选项
    \centering % 让盒子在页面上居中显示
    
    % 使用 \input 导入您的 tcolorbox 文件
    \begin{tcolorbox}[colback=orange!10!white, colframe=orange!80!black, 
  title=Reason All Mistakes Prompt, fonttitle=\bfseries]

You will be presented with the following content:\\
- A math problem:\\
\textbf{[Math Problem]}\\
\{problem\}\\
- A decomposed thought process for solving the problem:\\
\textbf{[Decomposed solution]}\\
\{decomposed solution in a list format\}\\

This thought process does not directly lead to the correct answer but includes attempts with incorrect reasoning, reflections, and corrections before arriving at the right solution.\\

- In Step \textless reflection start step index\textgreater \ of the thought process, a critical mistake occurs:\\
\textbf{[Earliest Step]}\\
\{reflection earliest step\}\\
- This mistake is identified through reflection in Step \textless reflection step index\textgreater:\\
\textbf{[Identified Fragment]}\\
\{reflection identified fragment\}\\

\textbf{**Task Description**}\\
- Your task is to identify the types of thinking steps within the range from step \textless reflection start step index \textgreater \ to step\textless reflection step index\textgreater \ that meet the following definitions. You need to check each definition in sequence. Once a definition is met, immediately label it with the corresponding step type and stop checking further.
\begin{description}
   \item[ - Reflection Steps:] A step that involves reflecting on or verifying the error and its resulting subsequent mistakes.
   \item[ - Incorrect Step:] A step is considered incorrect if: It contains mistakes itself or it continues the analysis based on the error from above steps without introducing a new, correct approach or making a proper correction.
\end{description}
\textbf{**Specific Requirements**}\\
- For each thinking step, carefully analyze and explain your reason before determining its type.\\
- The final output format should strictly adhere to standard \textbf{JSON}, for example:\\
\textasciigrave \textasciigrave \textasciigrave json
\begin{verbatim}
{"Step k": {"Analysis": "The specific
analysis for Step k before 
determining its type", 
"Step Type": ("Incorrect" / 
"Reflection" / "Correct")},...}
\end{verbatim}
\textasciigrave \textasciigrave \textasciigrave\\
\textbf{**Notes**}\\
- Ensure that your analysis of each step is well-reasoned and avoids subjective assumptions.\\

\end{tcolorbox}
    
    % --- 在这里添加标题和标签 ---
    \caption{The prompt used to determine the category of each step within the source-to-reflection range. This prompt utilizes information returned from the prompts in Table ~\ref{tab:aware_reflection_prompt} and Table ~\ref{tab:aware_mistake_source_prompt}.}
    \label{tab:reason_all_mistake_prompt} % 用于交叉引用的标签
\end{table}
\begin{table}[htbp] % htbp 是建议的浮动位置选项
    \centering % 让盒子在页面上居中显示
    
    % 使用 \input 导入您的 tcolorbox 文件
    \begin{tcolorbox}[colback=orange!10!white, colframe=orange!80!black, 
  title=Answer Decompose Prompt, fonttitle=\bfseries]

You will be presented with the following content:\\

\textbf{[Math Problem]}\\
\{problem\}\\

\textbf{[Answer]}\\
\{answer of the problem\}\\

\textbf{**Task Requirements**}\\
Insert `\textless step\textgreater' and `\textless/step\textgreater' into the original answer to break down the above answer into several atomic reasoning steps. Each atomic reasoning step belongs to only one of the following actions:\\

1. **Problem Analysis**: Analyze the original problem, organize the conditions, and deepen the understanding of the problem.\\
2. **Set Subtask Goals**: Clearly define the specific goal to be solved next, such as setting up an equation.\\
3. **Calculating or Solving**: Calculate a variable or solve an equation.\\
4. **Self-Reflection**: Reflect on potential mistakes and assess whether previous thoughts contain mistakes.\\
5. **Verify Correctness**: Validate whether the final answer or intermediate conclusions are correct.\\
6. **Error Correction**: Rectify identified mistakes.\\
7. **Switch Thinking**: Change to a different thought for solving the problem.\\
8. **Arrive at the Final Answer**: Conclude with the final answer.\\

\textbf{**Output Format**}\\
1. The breakdown of steps should not be too coarse; each step should belong to only one type of action.\\
2. Separate the reasoning steps by inserting `\textless step\textgreater' and `\textless/step\textgreater'.\\
3. Besides inserting the `\textless step\textgreater' and `\textless/step\textgreater' tags, do not change the original solution.\\
4. Do not alter the original solution except for inserting the `\textless step\textgreater' and `\textless/step\textgreater' tags; keep the original solution complete.\\
5. You only need to reply with the complete solution, without outputing any additional content.\\

\end{tcolorbox}
    
    % --- 在这里添加标题和标签 ---
    \caption{The prompt used to decompose answer. We used the same set of eight atomic logical actions as detailed in Table ~\ref{tab:solution_decompose_prompt}, with only minor modifications to the phrasing.}
    \label{tab:answer_decompose_prompt} % 用于交叉引用的标签
\end{table}

\end{document}